\PassOptionsToPackage{table}{xcolor}
\documentclass{article} 
\usepackage{iclr2027_conference,times}

\usepackage{amsmath,amsfonts,bm}

\def\eqref#1{equation~\ref{#1}}

\def\1{\bm{1}}

\DeclareMathAlphabet{\mathsfit}{\encodingdefault}{\sfdefault}{m}{sl}
\SetMathAlphabet{\mathsfit}{bold}{\encodingdefault}{\sfdefault}{bx}{n}

\usepackage[T1]{fontenc}
\usepackage[utf8]{inputenc}
\usepackage{booktabs}
\usepackage{placeins}   
\usepackage{tabularx}
\usepackage{array}
\usepackage{enumitem}
\usepackage{graphicx}
\usepackage[table]{xcolor}
\definecolor{shade}{gray}{0.90}
\definecolor{dim}{gray}{0.45}
\newcommand{\cond}[1]{\textsc{#1}}
\newcommand{\muted}[1]{\textcolor{dim}{#1}}
\usepackage{amsthm}
\newtheorem{proposition}{Proposition}[section]
\usepackage{hyperref}
\usepackage{url}
\hypersetup{colorlinks=true,linkcolor=blue!60!black,citecolor=blue!60!black,urlcolor=blue!60!black}

\usepackage{etoolbox}
\usepackage{needspace}   
\AtBeginEnvironment{table}{\setlength{\belowcaptionskip}{\abovecaptionskip}\setlength{\abovecaptionskip}{0pt}}

\title{Same Reward, Different Skills: \\ When Multimodal RL Learns to Look}

\author{%
Haocun Ye$^{1,2}$, Xinlong Jiang$^{2}$, Qile Chen$^{3}$ \\
Bingyu Wang$^{2}$, Teng Zhang$^{2}$, Shubai Chen$^{1,2}$ \\
Tingyu Wu$^{1,2}$, Zhenkun Zheng$^{1,2}$, Yiqiang Chen$^{2}$ \\[4pt]
{\normalfont\small $^{1}$University of the Chinese Academy of Sciences} \\
{\normalfont\small $^{2}$Institute of Computing Technology, Chinese Academy of Sciences} \\
{\normalfont\small $^{3}$Independent Researcher} \\[2pt]
{\normalfont\small Contact: \texttt{yehaocun23s@ict.ac.cn}}}

\iclrpreprintcopy
\begin{document}

\maketitle

\begin{abstract}
\looseness=-1 
Reinforcement learning with verifiable rewards (RLVR) improves vision-language benchmark scores even without visual information during training. With images at test, blind-trained models recover roughly half of the real-image gain at 3B and nearly four fifths at 7B. Prolonged real-image training can erode grounding while benchmark gains persist. Both findings expose the same gap: an image in the prompt is not an image in the learning signal. Our design rule, visual resolvability, asks that visual evidence be necessary for a correct answer and that the task remain learnable. We test it on counterfactual coordinate scenes in which the question stays fixed and the target is never named, so a correct answer requires finding the target in the image. With standard GRPO and correctness-and-format rewards, a 7B model raises its accuracy at finding the target (discovery) from 0.425 to 0.875 on held-out scenes denser than any it trained on, and it improves on question types it never trained on. Two controls locate the source of the gain. Replacing test images with gray canvases drops discovery to zero; training on gray canvases instead, at matched step 30 and in each of four seeds, yields essentially none of the gain even when the model is then tested with real images. The learned skill carries over to grounding tasks built independently of the training corpus. A caption that answers the training question, added to the same images, reward and budget, cuts the gain by nearly two thirds. Changing what reward requires changes what RL learns.

\end{abstract}

\section{Introduction}\label{sec:intro}

Reinforcement learning with verifiable rewards~\citep[RLVR;][]{lambert2025tulu3,guo2025deepseekr1} improves pretrained vision-language models by rewarding correct answers to image-based questions~\citep{huang2026visionr1,meng2025mmeureka,wang2025vlrethinker}. Gains on visual reasoning benchmarks~\citep{lu2024mathvista,zhang2024mathverse} are read as improvements in perception as well as reasoning. This interpretation guides data curation, reward design, and checkpoint selection. It rests on an assumption: rewarding correct answers develops the visual operations needed to produce them.

On Geometry3K~\citep{lu2021intergps}, replace every training image with a gray canvas, and the model still recovers roughly half of the real-image training gain at 3B and nearly four fifths at 7B when evaluated with real images. These are \textbf{blind gains}, learned without visual information during RL. Keeping the images is no guarantee either: prolonged real-image training with a trainable vision encoder can erode the ability to locate and read visual targets while benchmark gains persist---\textbf{grounding corrosion}~\citep[cf.][]{phan2026beyond}. Independent blind-trained runs lose many of the same previously solved examples, usually with identical wrong answers.

\begin{figure}[t]
\centering
\includegraphics[width=\linewidth]{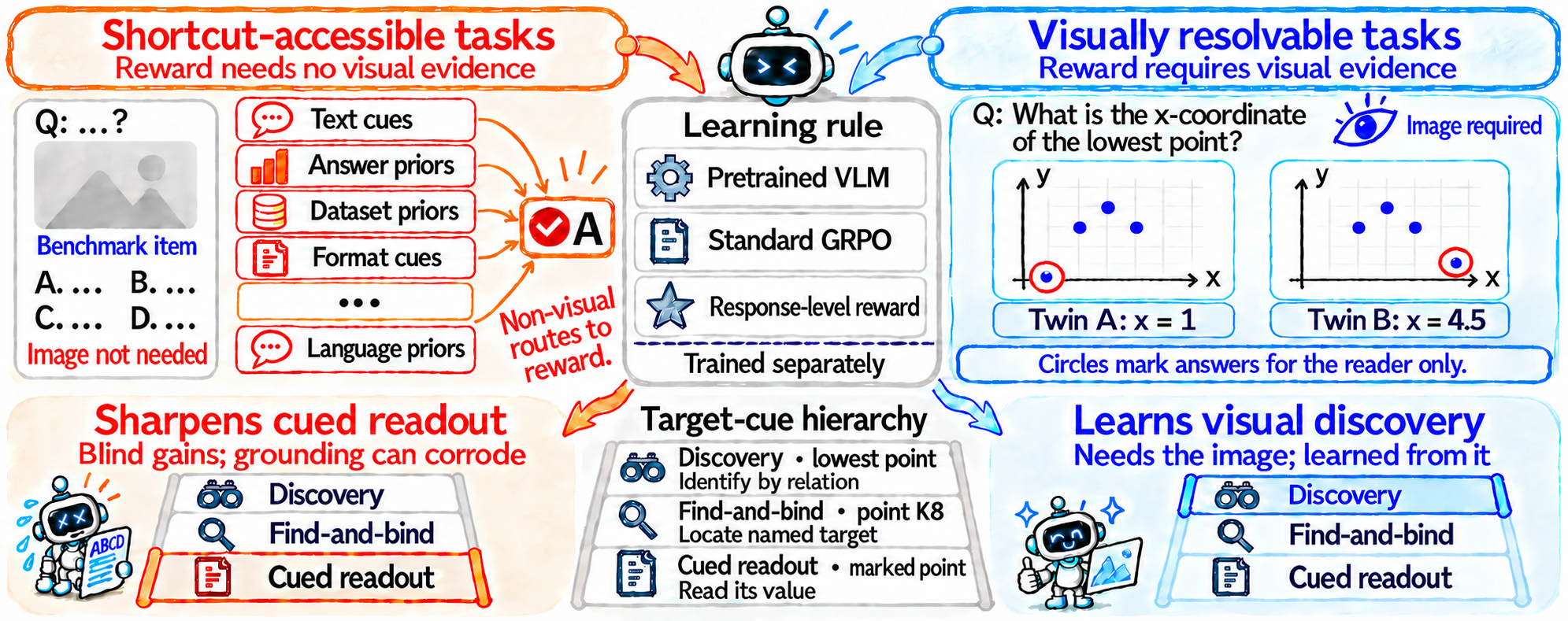}
\caption{\emph{Same learning rule, different lessons.} Left: shortcut-accessible tasks allow reward without visual evidence; gains in the reported 3B comparisons mainly favor cued readout. Right: constructed tasks demand visual evidence and relation-based target identification; discovery, cued readout, and named-target grounding improve. The hierarchy distinguishes supplied locations, supplied identities, and relation-defined targets. GRPO is shared across the study; rewards stay fixed within matched comparisons.}
\label{fig:overview}
\end{figure}

The design problem is therefore not just what the prompt contains, but what earning reward requires. \textbf{Shortcut-accessible} tasks permit correct answers without visual evidence. \textbf{Visual resolvability} is the alternative design rule: require visual evidence for correct answers while keeping the task learnable. The aim is to make the intended visual operation part of earning reward, rather than something the learner can bypass (Figure~\ref{fig:overview}\label{first:fig:overview}).

We put this rule into practice through two task controls: the visual fact and the target cue.\footnote{Code and data: \url{https://anonymous.4open.science/r/learning-without-looking-2432}} Counterfactual scenes keep the question fixed while changing one visual fact. Target cues range from a marked location or name to a relation such as ``the lowest point,'' which leaves the target to be identified (discovery). A base-model audit measures visual benefit and learnability. We train on relation-only coordinate questions and target-identification probes using standard GRPO~\citep{shao2024deepseekmath} with correctness-and-format rewards.

On a separate confirmatory set of 20-point scenes---a density absent from training---discovery accuracy rises from 0.425 to 0.875. Two visual-access controls separate needing images from learning through them: replacing test images with gray canvases reduces discovery accuracy to zero; at matched step 30, in each of four seeds, a blind-trained control evaluated with real images recovers essentially none of the discovery gain. The gains extend to question types never trained---cued readout and named-target grounding on the same scenes---and to independently constructed coordinate-grounding tasks. Mixing constructed with ordinary data turns the discovery gain up or down. The full study spans nine counterfactual task families, two public training corpora, seven audited benchmarks, and over thirty GRPO runs (Appendix~\ref{app:program}).

Whether RLVR develops capabilities or makes existing ones easier to elicit remains open~\citep{yue2025does,shao2026spurious}. We ask which visual operations improve and whether training images account for the gain. Related approaches modify the objective~\citep{wang2026papo}, response structure~\citep{xia2026visionaryr1}, available evidence~\citep{zheng2026deepeyes}, reward~\citep{xiao2026perceptionr1,yu2025perceptionr1,liu2025visualrft}, or data selection~\citep{yu2025dapo,wang2025thinklite}. We use standard GRPO and hold rewards fixed within each matched visual-access comparison; task construction specifies the evidence and operations that successful answering requires. Difficulty-based selection asks how hard a lesson is; resolvability also asks what evidence it requires. Appendix~\ref{app:related} gives the full related work.

\section{Benchmark Gains Are Not Visual Acquisition}\label{sec:gains}

\subsection{Learning Without Looking}\label{sec:blind}

We run 100 steps of GRPO from pretrained Qwen2.5-VL-3B-Instruct~\citep{bai2025qwen25vl} on filtered Geometry3K and evaluate on its held-out test set, with the vision encoder frozen and three seeds per condition (Appendix~\ref{app:training}). Four conditions differ only in what accompanies each question during RL: the real image (\cond{Real}), image-derived text (\cond{Caption}), no image (\cond{None}), or a uniform gray canvas (\cond{Gray}).

Table~\ref{tab:access}\label{first:tab:access} crosses training-time and test-time visual access. With real test images, \cond{Real} gains +0.238 over the base model, \cond{None} +0.137, and \cond{Gray} +0.125. Let $g_c$ be condition $c$'s gain under real-image evaluation; its recovery share is $g_c/g_{\mathrm{REAL}}$: 0.57 for \cond{None} and 0.52 for \cond{Gray}. Yet the gains depend on images at test: remove the test images and every condition gains little (+0.036 for \cond{Real}, +0.032 for \cond{Gray}). Blind gains also persist on the 497 questions for which the base model already produces an extractable answer: \cond{Gray} and \cond{None} recover 51\% and 54\% of \cond{Real}'s gain (Table~\ref{tab:extract}). At 7B, \cond{Gray} recovers 0.78 (95\% CI [0.64, 0.92]). An improvement can therefore depend on an image at test without having been learned from images.

\begin{table}[t]
\caption{\textit{Blind gains.} Training without visual information recovers more than half of \cond{Real}'s benchmark gain when real images are supplied at test; with the test image removed, no condition gains more than 0.05. Base row: accuracy; other cells: gain over that base under the same evaluation; shares: fraction of \cond{Real}'s gain (three-seed pooled). All cells use the same answer matcher.}
\label{tab:access}
\centering
\small
\setlength{\tabcolsep}{6pt}
\begin{tabular}{l c >{\columncolor{shade}}c cc}
\toprule
& \multicolumn{2}{c}{Evaluated with the real image} & \multicolumn{2}{c}{Test image removed} \\
\cmidrule(lr){2-3}\cmidrule(lr){4-5}
Trained with & gain over base & share of \cond{Real}'s gain & gray canvas & no image \\
\midrule
\textit{base model (accuracy)} & \textit{0.175} & --- & \muted{\textit{0.090}} & \muted{\textit{0.068}} \\
\cond{Real}    & $+$0.238 & 1 (reference)      & \muted{$+$0.019} & \muted{$+$0.036} \\
\cond{Caption} & $+$0.171 & 0.72  & \muted{$+$0.024} & \muted{$+$0.037} \\
\addlinespace[3pt]
\multicolumn{5}{l}{\textit{trained without visual information}} \\
\cond{None}    & $+$0.137 & \textbf{0.57} & \muted{$+$0.015} & \muted{$+$0.044} \\
\cond{Gray}    & $+$0.125 & \textbf{0.52} & \muted{$+$0.014} & \muted{$+$0.032} \\
\bottomrule
\addlinespace[2pt]
\multicolumn{5}{l}{\footnotesize At 7B, \cond{Gray} recovers 0.78 [0.64, 0.92] of \cond{Real}'s gain (Table~\ref{tab:geo7b}).}
\end{tabular}
\end{table}

\subsection{How Much Does the Image Matter?}\label{sec:necessity}

Raw blind-to-image accuracy ratios credit guessing and can blur differences in visual dependence. We audit the base 3B model using chance-corrected blind retention, $\rho_{\mathrm{blind}}={(a_{\mathrm{blind}}-c)/(a_{\mathrm{image}}-c)}$ for $a_{\mathrm{image}}>c$, where $c$ is the chance reference: the mean of $1/k$ over multiple-choice items with $k$ options, and zero for free-form answers. BLINK~\citep{fu2024blink} and the multiple-choice portion of MathVerse~\citep{zhang2024mathverse} look alike under the raw ratio (0.83 and 0.85), but separate to 0.27 and 0.66 after correction. Audits of two other model families show a sharp contrast: without images, Gemma-3~\citep{gemma2025gemma3} keeps about three quarters of its free-form accuracy on ordinary questions from the RLVR corpus ViRL39K~\citep{wang2025vlrethinker}, and InternVL3-9B~\citep{zhu2025internvl3} about half, yet both score exactly zero on the grounding task below (Appendix~\ref{app:audit}). The audit asks whether the image matters; to see what RL changes, we separate the operations used to answer.

\subsection{Reading a Value, Finding a Target}\label{sec:ladder}

We distinguish tasks by how much target information is supplied. \textbf{Cued readout} marks the target's location and asks for its value. \textbf{Find-and-bind} names the target---a point in a coordinate plot---and requires locating it and reading its coordinate. \textbf{Discovery} supplies only a relation, such as ``the point with the smallest y-coordinate,'' and requires identifying the target before reading it. The cue hierarchy orders the information supplied, not model accuracy.

We measure cued readout and find-and-bind on a counterfactual suite: a \emph{grounding task} of 600 named-point coordinate pairs, a \emph{cued-readout task} of 300 marked-plot pairs, and a \emph{header-cued table task} of 300 pairs. Each pair keeps the question fixed and changes one visual fact so that its two correct answers differ. For a predictor $h$, let $C_{j,s}(h)\in\{0,1\}$ indicate correctness on member $s\in\{a,b\}$ of pair $j$. Pair accuracy~\citep{tong2024eyes,li2024naturalbench} over $N_{\mathrm{pair}}$ pairs is
\begin{equation}
\operatorname{PairAcc}(h)
=\frac{1}{N_{\mathrm{pair}}}\sum_{j=1}^{N_{\mathrm{pair}}}
 C_{j,a}(h)\,C_{j,b}(h).
\label{eq:pairacc}
\end{equation}
An image-independent greedy predictor returns the same extracted answer on both members. Their accepted answers are disjoint, so every product in Eq.~(\ref{eq:pairacc}) is zero (Proposition~\ref{prop:pairzero}). The grounding task has an independently regenerated twin, and human review confirmed all 60 audited pairs of the suite (Appendix~\ref{app:grounding}).

\begin{figure}[t]
\centering
\includegraphics[width=\linewidth]{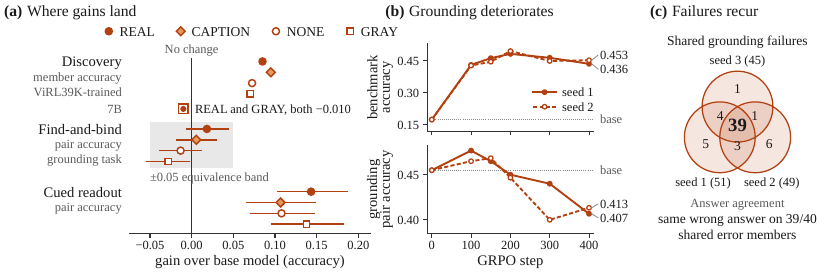}
\caption{\emph{Benchmark gains are not visual acquisition.} (a) Gains by operation: readout and grounding for Geometry3K-trained 3B models (three-seed means), discovery for ViRL39K-trained models (two-seed means; 7B: Geometry3K-trained pair). (b) Benchmark accuracy and grounding pair accuracy under prolonged \cond{Real} training with a trainable encoder. (c) Grounding pairs lost by three \cond{Gray}-trained runs, and their wrong answers. All evaluations use real images; error bars are 95\% intervals and shading marks the grounding equivalence band.}
\label{fig:necessity}
\end{figure}

On this suite, Geometry3K training improves cued readout under every visual-access condition: \cond{Real} gains +0.143, \cond{Gray} +0.138, and \cond{None} +0.108 (Figure~\ref{fig:necessity}\label{first:fig:necessity}(a)). Cued readout accounts for 72\% of \cond{Real}'s total pair-count-weighted improvement across the 1,200 pairs. On the grounding task, \cond{Real}'s gain of +0.018 is equivalent to zero within the pre-specified $\pm0.05$ band (90\% CI [$-0.003$, +0.040]; Appendix~\ref{app:stats}), and \cond{Gray} declines by 0.028. Reading a marked value improves; locating a named target shows little change.

Blind gains also reach discovery. For 3B models trained under the four visual-access conditions on ViRL39K, discovery gains range from about +0.04 to +0.10, and \cond{None} and \cond{Gray} finish within 0.04 of \cond{Real} at every density (two-seed means). At 7B, Geometry3K training barely changes 20-point discovery (0.470 $\rightarrow$ 0.460 under \cond{Real} and \cond{Gray}). In these ordinary 3B runs, gains mainly favor cued readout, and blind training reproduces most of the discovery gain.

\section{When Gains Conceal Grounding Loss}\label{sec:corrosion}

Under the long-horizon real-image recipe below, grounding does more than stall: it falls below the base model's while benchmark gains persist.

\textbf{Long-horizon training.}\label{sec:decay} We train Qwen2.5-VL-3B-Instruct for 400 GRPO steps on unfiltered Geometry3K, with real images throughout, a trainable vision encoder, and correctness-and-format rewards, in two independent runs evaluated at step 400. Benchmark accuracy rises from 0.175 to 0.436 and 0.453, while grounding-task pair accuracy, after a small rise, falls by 0.070 and 0.052 from step 100 (95\% CIs exclude zero) to 0.407 and 0.413, below the base model's 0.455 from step 200 onward (Figure~\ref{fig:necessity}(b)). The first run's near-flat benchmark score from step 100 to 400 hides item-level churn (Appendix~\ref{app:corrosion}). We call this loss of previously available visual grounding under reward optimization \emph{grounding corrosion}.

\textbf{Recurring errors.}\label{sec:same-items} In a separate analysis, the three \cond{Gray}-trained 3B runs from \S\ref{sec:blind}, evaluated with real images, each lose 45--51 grounding-task pairs that the base model solved. Their degraded sets overlap with a Jaccard index of 0.661, against a permutation-null mean of 0.011 for sets of the same sizes (Figure~\ref{fig:necessity}(c)). Among the degraded pairs, 40 members are wrong in all three runs, with identical answers on 39: the runs agree not only on what fails, but on what replaces the correct answer. Those 39 answers span 11 distinct values. On 29 of these 39 members, at least one \cond{Real}-trained run gives the same wrong answer: these item-structured errors recur across visual-access conditions. Whatever produces them, it is not the training images.

\textbf{Grounding at 7B.}\label{sec:scale} The larger model shows a different balance: real-image training improves grounding even though much of the benchmark gain remains available to blind training. In the ordinary 7B access comparison, \cond{Real} gains +0.035 (95\% CI [0.013, 0.058]) on the grounding task and +0.038 on its twin. \cond{Gray}'s changes (+0.012 and +0.013) have intervals that include zero, as do both models' changes on cued readout (Table~\ref{tab:access7b}).

\section{The Missing Link: What Reward Requires}\label{sec:link}

Images help at test. The puzzle is whether earning reward makes the model learn from them. Our account is that optimization can strengthen nonvisual answer strategies when these suffice to earn reward; the resulting policy may still exploit images supplied at test. Such a policy has not failed to look; nothing required it to. A target-location cue offers another shortcut: it removes search while leaving the value to be read from the image. Needing the image is not the same as needing to discover what to read.

Two properties guide task construction. \textbf{Visual necessity} ties correct answers to visual evidence rather than an image-free shortcut. \textbf{Learnability} leaves room for informative correctness feedback: when responses are almost uniformly right or wrong, there is little variation in correctness to guide learning. Together, these properties define a \textbf{visually resolvable} training problem. The \emph{Visual Resolvability Principle} is a design rule: build learnable tasks whose successful solutions require the visual operation we want to develop. We instantiate it in counterfactual coordinate scenes, where both properties can be measured and controlled directly.

\section{Make Looking Necessary: Measure, Shape, Verify}\label{sec:make}

Two controls specify what the task demands: which visual fact determines the answer, and what the prompt reveals about its target. A base-model audit measures the contribution of the image and the opportunity for correctness feedback.

\subsection{Measure: Visual Necessity and Learnability}\label{sec:measure}

We audit the base Qwen2.5-VL-7B-Instruct model before RL: 16 responses per item with its image and 16 without, at temperature 1, scored by the training reward's accuracy component. Let $b_i$ and $q_{\mathrm{blind},i}$ be these empirical success rates. Our empirical visual-necessity measure, $\Delta q_i=b_i-q_{\mathrm{blind},i}$, estimates how much the image raises correct-answer probability; the corpus mean of $q_{\mathrm{blind},i}$ is its \emph{blind reward opportunity}.

The audited learnability band, $0<b_i<1$, means that both correct and incorrect responses were observed. At a fixed prompt and checkpoint, let $p_i$ be the true correctness probability of an independently sampled response. A group of $G=5$ responses contains both outcomes with probability $P_{\mathrm{mix}}(p_i;G)=1-p_i^G-(1-p_i)^G$; the subtracted terms are the all-correct and all-incorrect cases. The band thus targets opportunities for accuracy-based reward variation (Appendix~\ref{app:feedback}). This fixed audit describes the corpus and supplies the mixture measurements; it neither filters nor reweights training. The constructed corpus has blind reward opportunity 0.003, and 92.9\% of its rows satisfy $\Delta q_i>0$ and $0<b_i<1$ (Figure~\ref{fig:design}\label{first:fig:design}(a)).

\subsection{Shape: Change the Fact, Change the Cue}\label{sec:shape}

Each scene program specifies labeled points, their coordinates, and the relations from which questions and exact answers derive. Rendering a program's two twins changes one answer-relevant fact while the question and the rest of the scene stay fixed.

\begin{figure}[t]
\centering
\includegraphics[width=\linewidth]{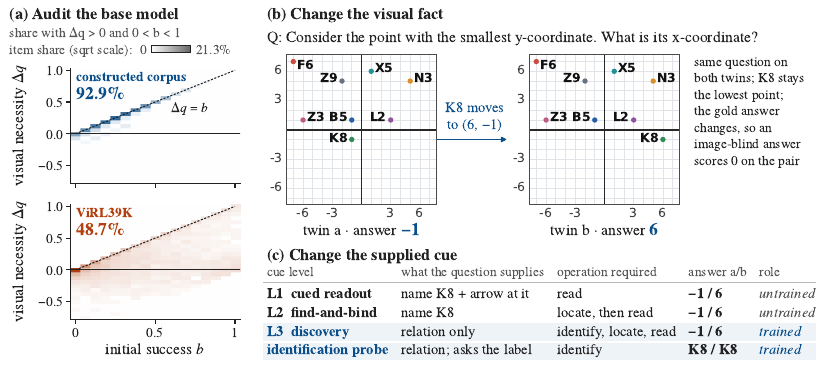}
\caption{\emph{Measure and shape.} (a) Audited visual benefit against initial success, constructed corpus and ViRL39K. (b) The K8 scene and its counterfactual twin. (c) Its target-cue hierarchy.}
\label{fig:design}
\end{figure}

\textbf{Changing the visual fact.} In the eight-point scene of Figure~\ref{fig:design}(b), the question asks, ``Consider the point with the smallest y-coordinate. What is its x-coordinate?'' The target, K8, lies at $(-1,-1)$. Its counterfactual twin moves K8 to $(6,-1)$: K8 remains the lowest point, but the answer changes from $-1$ to $6$. The question is unchanged, and solving the pair requires both answers to be correct (Eq.~(\ref{eq:pairacc})).

\textbf{Changing the supplied cue.} On the same scene, cued readout (L1) names K8 and marks it with an offset arrow; find-and-bind (L2) names K8 without the arrow; discovery (L3) gives only the relation. All three ask for the same coordinate; only the supplied target information changes. Discovery requires identifying the relevant point before reading its coordinate. A separate \textbf{identification probe} asks for the label directly: ``Which labeled point has the smallest y-coordinate?''

The corpus balances three pair roles: \emph{target-stable} pairs move the target, as in the K8 scene; \emph{target-switch} pairs change which point satisfies the relation; and \emph{invariance} pairs change the image without changing the answer. Answers must follow relevant changes and ignore irrelevant ones. From each of 720 scene programs we take the discovery question and probe on both twins, giving 2,880 training prompts on 8- and 12-point scenes. The 20-point scenes are reserved for evaluation; L1 and L2 are evaluated but never trained. Discovery accuracy uses target-stable and invariance pairs, with target-switch pairs reported separately.

\subsection{Verify: Does the Test Require Looking?}\label{sec:verify}

We track learning on a fixed development set and assess acquisition on a separate confirmatory set. Five criteria qualify each scene density: discovery neither trivial nor unreachable for the base model; discovery harder than both cued levels; zero image-removed accuracy on discovery and its probe; question-blind captions insufficient to answer; and failure of the specified linear classifiers to distinguish the twins. We set the numerical thresholds before generating 450 new scene programs, 150 per density, disjoint from training and development. The 20-point subset meets all five and supplies the confirmatory result; Appendix~\ref{app:ladder} reports every density's outcome on both sets and the protocol timing.

\subsection{Train: Standard GRPO}\label{sec:grpo}

We train the 7B base model with standard GRPO. The vision encoder and projector stay frozen; only the language model is trained. For each image--question prompt we sample five responses at temperature 1 and score each response $o$ independently against the gold answer $y$:
\begin{equation}
r(o,y)=\frac{1}{2}\,\mathbf{1}\{\operatorname{correct}(o,y)\}
      +\frac{1}{2}\,\mathbf{1}\{\operatorname{format}(o)\}.
\label{eq:reward}
\end{equation}
Correctness tests the extracted answer for equivalence to $y$; format requires exactly one nonempty \texttt{<answer>\ldots</answer>} span. GRPO normalizes rewards within each prompt's five responses. This reward stays fixed across the constructed-corpus visual-access and mixture comparisons. Twins are separate training rows with separate response groups: pairing shapes the data; rewards remain response-level. Training spans 100 steps in four segments; restarts reset AdamW moments, re-anchor the KL reference, and restart the corpus order. All matched 30-step comparisons end before the first restart.

\section{The Regime Flips}\label{sec:flip}

Training on relation-only questions and identification probes improves discovery and the cued tasks on the same held-out scenes.

\subsection{Learning to Find What to Read}\label{sec:acquired}

On the 20-point confirmatory set, discovery accuracy rises from 0.425 to 0.740 after 30 steps and to 0.875 after 100 (in four segments; \S\ref{sec:grpo}), a gain of 0.450 (95\% CI [0.365, 0.540]; Table~\ref{tab:confirmatory}). The same scenes also show gains on the cued tasks, which were never trained: at step 100, cued readout rises from 0.610 to 0.890 and find-and-bind from 0.620 to 0.950. At matched step 30, the two cued tasks gain +0.210 and +0.280, against +0.030 and +0.060 for \cond{Gray}. On target-switch pairs, discovery improves from 43 to 78 correct answers out of 100 at step 30, against 44 for \cond{Gray}, and to 88 at step 100 (Table~\ref{tab:hierarchy}).

\begin{figure}[t]
\centering
\includegraphics[width=\linewidth]{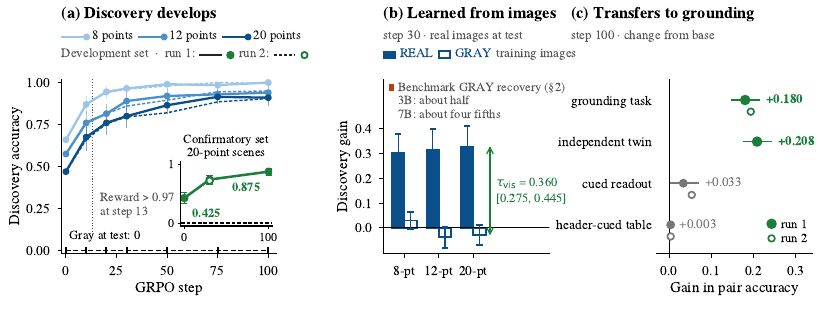}
\caption{\emph{The regime flips.} (a) Discovery member accuracy on the development set (inset: confirmatory 20-point result). (b) Step-30 discovery gains under \cond{Real} and \cond{Gray} training, evaluated with real images. (c) Step-100 change in pair accuracy. Error bars: 95\% intervals over scene programs in (a, b) and over pairs in (c).}
\label{fig:flip}
\end{figure}

Learning continues long after the reward saturates: training reward exceeds 0.97 by step 13, while discovery on the 12- and 20-point development scenes rises through step 75 (Figure~\ref{fig:flip}\label{first:fig:flip}(a)). The result replicates: a second run of the same configuration ends within 0.010 of the first at every development density (Table~\ref{tab:devgrid}), and three independent seeds reach 0.730--0.760 on the confirmatory set at step 30; seed 1 reaches 0.895 at step 100 (Table~\ref{tab:seeds}).

The gain does not hinge on the relation-only question. Training instead on the named-target question (find-and-bind) with the same probes, three seeds each, leaves confirmatory discovery +0.042 below the relation-only corpus (seed-level 95\% interval [$-$0.027, +0.110]; Table~\ref{tab:seedcontrasts}): identification and grounding, trained separately, compose into discovery. Identification is itself learned: on members the base model could not identify, \cond{Real} lifts discovery from 0.143 to 0.629, against 0.485 to 0.764 where it could; \cond{Gray} improves neither (Table~\ref{tab:composite}).

\subsection{Using Images, Learning from Images}\label{sec:drives}

One control tests what the model uses; the other tests what taught it.

\textbf{Images at test.} Replacing test images with gray canvases reduces discovery and probe accuracy to exactly zero for both \cond{Real} runs at every development checkpoint and the evaluated confirmatory checkpoints. With the image gone, the trained models answer like the untrained one, with default responses outside the answer set (Appendix~\ref{app:acquisition}).

\textbf{Images during training.} \cond{Gray} uses the same 2,880 prompts, starting checkpoint, reward, and budget, but replaces every training image with a gray canvas: it is shown everything \cond{Real} is shown, except the image. We evaluate both models with real images at matched step 30. The visually attributable gain $\tau_{\mathrm{vis}}(t)$ is \cond{Real}'s gain minus \cond{Gray}'s over the common image-present base (Appendix~\ref{app:measures}). On development scenes, \cond{Real} gains 0.305--0.330 across densities, \cond{Gray} changes by $-0.035$ to +0.030, and $\tau_{\mathrm{vis}}(30)=0.360$ (95\% CI [0.275, 0.445]) on the never-trained 20-point density (Figure~\ref{fig:flip}(b)); the confirmatory set shows the same contrast (Table~\ref{tab:attribution}). The blind-gain pattern reverses: on Geometry3K, blind training recovered much of the real-image gain; here it recovers essentially none. The reversal holds seed by seed: across four independent seeds, \cond{Real} exceeds \cond{Gray} by +0.354 on the confirmatory set (seed SD 0.019) and by +0.339 on an untouched set of 1,332 members (Table~\ref{tab:seedcontrasts}).

\textbf{Answers without looking.} \cond{Gray} removes the image; a third arm keeps it and adds a shortcut: the question-blind caption of each training image goes into the prompt, with the same scenes, reward, budget and seeds, so that at the training densities the answer can be read from text. Reward is earned as before (accuracy component 0.97--0.99 by step 30); the skill is not. With the image alone, this arm gains +0.120 on confirmatory discovery against \cond{Real}'s +0.323, below \cond{Real} in every seed (gap +0.203, seed-level 95\% interval [+0.163, +0.243]); even with its caption at test it reaches at most 0.655 (Table~\ref{tab:shortcut}). Same reward, different skill.

\subsection{Discovery Training Transfers to Coordinate Grounding}\label{sec:transfer}

Training on discovery and identification also improves find-and-bind on the grounding task and its independently regenerated twin. Both predate the training corpus and were never used in training; they were built to catch shortcuts, not to be trained on. At step 100, grounding-task pair accuracy rises from 0.768 to 0.948 (+0.180, 95\% CI [0.147, 0.215]) and twin pair accuracy from 0.728 to 0.937. The second run reaches 0.962 on the grounding task, and seed 1 at step 100 reaches 0.942. The cued-readout and header-cued table tasks, built in other visual formats, change by at most 0.053 (Figure~\ref{fig:flip}(c); Table~\ref{tab:transfer}). At matched step 30, \cond{Real} gains +0.162 on grounding and the \cond{Gray}-trained control +0.002: the transfer to coordinate grounding comes from training images. Across the four seeds, \cond{Real} exceeds \cond{Gray} on the grounding task by +0.141 (seed SD 0.029; Table~\ref{tab:seedcontrasts}).

\section{Turning the Dial}\label{sec:dial}

\subsection{Changing the Mixture, Changing the Gain}\label{sec:dose}

We mix constructed data with ordinary ViRL39K examples while holding the training budget fixed. In each 240-prompt step, a fraction $f\in\{0,\tfrac13,\tfrac23,1\}$ is constructed and the rest ordinary. Every condition starts from the same 7B model, keeps the reward and recipe, and sees 7,200 prompt presentations in 30 steps. We manipulate the constructed fraction $f$; the base-model audit measures the resulting stream's resolvability,
\begin{equation}
m(f)=f\,\overline{\Delta q}_{\mathrm{constructed}}
 +(1-f)\,\overline{\Delta q}_{\mathrm{ViRL},f}.
\label{eq:mass}
\end{equation}
Here $m$ measures a correct-answer probability difference, not a data fraction, and spans 0.126 to 0.249. Each grid point is a single run at the same budget.

\begin{figure}[t]
\centering
\includegraphics[width=\linewidth]{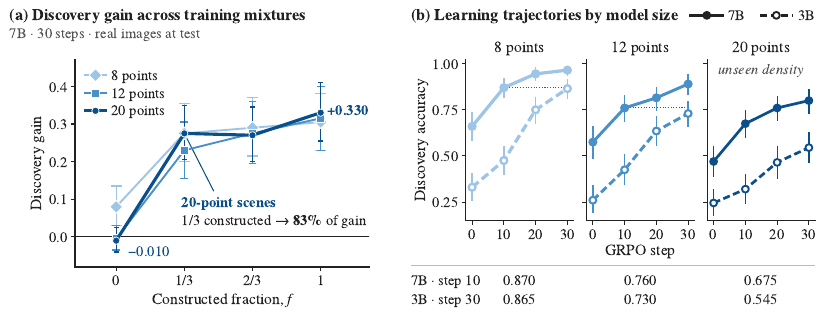}
\caption{\emph{Composition changes the gain; model size changes the trajectory.} (a) Step-30 discovery gain of each training mixture against the constructed fraction $f$. (b) Discovery accuracy of 3B and 7B models trained on the constructed corpus. Error bars: 95\% intervals over scene programs.}
\label{fig:dial}
\end{figure}

On the never-trained 20-point density, ordinary data alone leaves discovery where it was ($-0.010$), a one-third constructed mixture gains +0.275, and the full corpus gains +0.330 (Figure~\ref{fig:dial}\label{first:fig:dial}(a)). One-third constructed data thus delivers over 80\% of the full-corpus gain at the same budget. The full corpus gives the largest gain at every density (Table~\ref{tab:dose}). Paired comparisons on the same scenes favor the full corpus over either mixture by about 0.06, and the two mixtures do not differ (Table~\ref{tab:slopes}).

The one-third mixture keeps ordinary-task accuracy. With images, it matches the untrained model on 4,239 held-out ViRL39K items (0.519 against 0.509), MathVista testmini (0.696 against 0.682) and BLINK (0.547 against 0.536); ordinary-only training reaches 0.548 on ViRL39K and the fully constructed run 0.419, its deficit concentrated on multiple-choice items (Appendix~\ref{app:dose}).

At equal constructed exposure (2,400 presentations) the one-third mixture is ahead of the fully constructed run, 0.745 against 0.675 (Appendix~\ref{app:dose}).

\subsection{Same Tasks, Different Learning Curves}\label{sec:capacity}

The smaller model improves too, but on a different timetable. We train Qwen2.5-VL-3B on the same constructed corpus with the same recipe and 30-step budget; its identification-probe accuracy remains zero under gray-canvas evaluation (Table~\ref{tab:capacity}). On held-out scenes at the training densities, 3B at step 30 roughly matches 7B at step 10: 0.865 against 0.870, and 0.730 against 0.760 (Figure~\ref{fig:dial}(b)). On the never-trained density it reaches 0.545, against 0.675. Grounding-task performance improves at both scales (+0.130 at 3B, +0.162 at 7B; step 30).

\section{Discussion}\label{sec:discussion}

Specifying the reward does not finish specifying the learning problem. The verifier credits answers; the task determines what producing those answers requires. A correctness reward can reinforce an answer-producing shortcut or support visual discovery. In the mixture experiment, changing composition changes acquisition while model, reward, optimizer, and budget stay fixed. Task design, not only reward design, shapes the lesson.

\textbf{Score is not skill.} A higher score can conceal where a gain came from and which skill was lost along the way. Benchmark improvement, dependence on visual input, and learning from visual evidence are distinct claims. Image removal tests whether successful answers depend on the image; matched blind training tests whether the improvement was learned from it. A claim of visual learning must name the operation acquired and the information that taught it.

\textbf{The learning problem is the lever.} Reward engineering and data selection are means; the practical question is whether the learning problem makes the intended operation necessary for success. Data construction is one way to write that requirement into the task; the named-target ablation shows it need not take one question form (\S\ref{sec:acquired}). The operation should also guide evaluation: benchmark gains can persist while grounding erodes, and discovery keeps improving after reward saturates. Checkpoint selection should track the skill being taught, not reward or aggregate score alone.

\textbf{Scope.} Coordinate scenes are a deliberate testbed, in which visual necessity can be verified directly and the target cue varied while the question stays fixed. Within it, training on relation-defined discovery and identification also improves question types that were never trained: cued readout and named-target grounding at a held-out scene density, and grounding tasks built before the corpus. The discovery gain appears in every run trained on real constructed images, at both model sizes. Applying the principle to other visual formats and to natural images, where visual necessity must first be measured, is the next test of its generality.

\textbf{Beyond vision.} In the constructed-corpus runs, the vision encoder and projector stay frozen; training changes only the language model. A given image supplies the same visual tokens before and after, yet the model becomes better at identifying a target, grounding it, and reading its value. Learning to see, in these experiments, was learning to look. The broader proposal concerns evidence, not pixels: an audio task can demand an acoustic distinction that a transcript omits; a video task can require temporal order; a tool-use task can make the returned result determine the answer. In each, the goal is to tie verifiable success to a chosen source of evidence while keeping the operation learnable. What matters is not how many modalities a prompt contains but which evidence the model must learn to use to succeed. That turns the question around: instead of observing which capabilities emerge, we design the learning problems that develop them.

The verifier decides what counts; the data determine what it takes. Changing what reward requires changes what RL learns.\label{end:main}

\needspace{4\baselineskip}
\subsection*{AI use statement}
AI tools were used solely to polish the language and improve the readability of the manuscript. The authors conducted the
literature review, developed the research ideas, carried out the experiments, selected the references, and wrote the initial
draft. All AI-assisted edits were reviewed by the authors, who take full responsibility for the final content.

\subsection*{Reproducibility statement}
Anonymized code: \url{https://anonymous.4open.science/r/learning-without-looking-2432}. The repository holds the scene
program and the counterfactual grounding suite with its regenerated twin (Appendices~\ref{app:ladder} and~\ref{app:grounding}),
corpus preparation and the training mixtures of \S\ref{sec:dose}, the base-model audit (\S\ref{sec:measure},
Appendix~\ref{app:audit}), the training reward and answer matcher (\S\ref{sec:grpo}, Appendix~\ref{app:stats}), every
training recipe with the modifications to the training framework (Appendix~\ref{app:training}), and the evaluation and
statistics code (Appendices~\ref{app:measures} and~\ref{app:stats}). From the released per-item evaluation outputs, training
streams, and audit rows, one script rebuilds the paper's tables and figure values without a GPU and compares each with the
printed value; six printed values, such as the training-reward curve, come from trainer logs and are not rebuilt. Re-running
training and evaluation requires GPUs; the repository's scripts fetch the released scene sets, grounding suite, corpora, and
the 7B model trained on the constructed corpus.

\bibliography{references}
\bibliographystyle{iclr2027_conference}

\setcounter{topnumber}{4}
\setcounter{totalnumber}{6}
\makeatletter
\newcount\app@barrier@n
\newcommand{\AppendixBarrier}{\par\begingroup \let\@elt\relax
  \edef\@tempa{\@botlist\@deferlist\@dbldeferlist}%
  \ifx\@tempa\@empty
  \else
    \ifnum\app@barrier@n>3 \clearpage            
    \else \global\advance\app@barrier@n 1 \null\newpage \AppendixBarrier \fi
  \fi \endgroup
  \global\app@barrier@n=0 \suppressfloats[t]}
\makeatother
\appendix
\counterwithin{table}{section}
\counterwithin{figure}{section}
\counterwithin{equation}{section}

\AppendixBarrier
\section{Visual Necessity Audit}\label{app:audit}
For each benchmark we compare accuracy with the image against accuracy with the image removed. For the retention $\rho_{\mathrm{blind}}$ of \S\ref{sec:necessity}, the chance reference is $c=N^{-1}\sum_i 1/k_i$ for a multiple-choice task with $k_i$ options on item $i$ and $c=0$ for free-form answers; mixed-format benchmarks are reported by format. Retention is a ratio of aggregate above-chance accuracies, not an average of per-item ratios; it is defined when $a_{\mathrm{image}}>c$, need not lie in $[0,1]$, and is not clipped. Table~\ref{tab:audit} audits the untrained 3B model on seven benchmarks: BLINK, MathVerse, MMVP, MMMU~\citep{yue2024mmmu}, HallusionBench~\citep{guan2024hallusionbench}, MMStar and MathVista. Table~\ref{tab:crossfamily} repeats the audit for two model families we did not train, Gemma-3 and InternVL3-9B; these are additional base-model audits, not tests of the intervention on other families.
\begin{table}[h]
\caption{Visual-necessity audit of the untrained Qwen2.5-VL-3B model on seven benchmarks: accuracy with the image and with the image removed (blind), the chance level (mean of the item-specific $1/k$ for multiple choice; 0 for free-form answers), naive retention (blind $\div$ with-image) and chance-corrected retention $(\text{blind}-\text{chance})/(\text{with-image}-\text{chance})$, with 95\% item-paired bootstrap intervals (10,000 draws). Mixed-format benchmarks are reported by format; HallusionBench is reported under both plausible chance levels.}
\label{tab:audit}
\centering
\scriptsize
\setlength{\tabcolsep}{3pt}
\begin{tabular}{l r r r r l l}
\toprule
benchmark (format) & $n$ & with image & blind & chance & naive retention & corrected retention \\
\midrule
BLINK (MC, pooled) & 1901 & 0.493 & 0.409 & 0.377 & 0.829 [0.781, 0.879] & 0.271 [0.089, 0.454] \\
MathVerse (MC, pooled) & 2180 & 0.465 & 0.394 & 0.260 & 0.848 [0.807, 0.889] & 0.655 [0.568, 0.743] \\
MathVerse (free-form) & 1755 & 0.055 & 0.019 & 0.000 & 0.340 [0.240, 0.458] & 0.340 [0.240, 0.458] \\
MMVP (MC, $k=2$) & 300 & 0.660 & 0.500 & 0.500 & 0.758 [0.668, 0.854] & 0.000 [$-$0.417, 0.342] \\
MMMU dev+val (MC, pooled) & 988 & 0.506 & 0.413 & 0.263 & 0.816 [0.764, 0.873] & 0.617 [0.512, 0.729] \\
MMMU dev+val (free-form) & 62 & 0.097 & 0.048 & 0.000 & 0.500 [0.000, 1.000] & 0.500 [0.000, 1.000] \\
HallusionBench (free-form, chance 0) & 1129 & 0.598 & 0.475 & 0.000 & 0.794 [0.748, 0.843] & 0.794 [0.748, 0.843] \\
HallusionBench (Yes/No, chance 0.5) & 1129 & 0.598 & 0.475 & 0.500 & 0.794 [0.748, 0.843] & $-$0.258 [$-$0.673, 0.039] \\
MMStar (MC, pooled) & 1500 & 0.554 & 0.261 & 0.269 & 0.471 [0.430, 0.513] & $-$0.029 [$-$0.108, 0.049] \\
MathVista testmini (MC, pooled) & 539 & 0.725 & 0.512 & 0.332 & 0.706 [0.646, 0.765] & 0.458 [0.351, 0.564] \\
MathVista testmini (free-form) & 460 & 0.504 & 0.115 & 0.000 & 0.228 [0.174, 0.287] & 0.228 [0.174, 0.287] \\
\bottomrule
\end{tabular}
\end{table}

\begin{table}[h]
\caption{The same audit for two other model families on a 4,096-item ViRL39K sample (the image is removed rather than replaced): with-image and blind accuracy, the chance level, and naive and chance-corrected retention with 95\% item-paired bootstrap intervals. Corrected retention is defined only when with-image accuracy exceeds the chance level (\S\ref{sec:necessity}). On the multiple-choice pool Gemma-3 scores below chance with the image, so its ratio is undefined; InternVL3-9B exceeds chance by only 0.026, so its ratio is defined but unstable and is reported unclipped. Without the image, both models score 0.000 pair accuracy on the grounding task and its twin (\S\ref{sec:necessity}).}
\label{tab:crossfamily}
\centering
\scriptsize
\setlength{\tabcolsep}{3pt}
\begin{tabular}{l l r r r r l l}
\toprule
model & pool & $n$ & with image & blind & chance & naive retention & corrected retention \\
\midrule
Gemma-3 & ViRL39K free-form & 2789 & 0.430 & 0.312 & 0.000 & 0.727 [0.690, 0.765] & 0.727 [0.690, 0.765] \\
Gemma-3 & ViRL39K multiple choice & 1215 & 0.135 & 0.086 & 0.268 & 0.634 [0.533, 0.744] & undefined \\
InternVL3-9B & ViRL39K free-form & 2789 & 0.269 & 0.130 & 0.000 & 0.485 [0.439, 0.533] & 0.485 [0.439, 0.533] \\
InternVL3-9B & ViRL39K multiple choice & 1215 & 0.294 & 0.205 & 0.268 & 0.697 [0.619, 0.782] & $-$2.439 [$-$17.961, 6.505] \\
\bottomrule
\end{tabular}
\par\vspace{2pt}
{\scriptsize\raggedright The two pools hold 4,004 of the 4,096 sampled items; 92 multiple-choice items whose number of options could not be determined are excluded.\par}
\end{table}

\AppendixBarrier
\section{Task Construction}\label{app:ladder}
\textbf{Construction.} Each scene program yields two twin images that differ in one fact, and three question levels on the same scene (Table~\ref{tab:ladder}): L1 marks the target's position, L2 names the target, and L3 gives only the relation; a probe asks which point satisfies the relation. Pairs have one of three roles, in equal numbers: \emph{target-stable} pairs move the target, so the answer changes while the same point satisfies the relation (the K8 scene); \emph{target-switch} pairs change which point satisfies it; and \emph{invariance} pairs change the image without changing the answer.

\textbf{Scoring.} Discovery accuracy is member accuracy over the target-stable and invariance pairs, 200 questions from 100 scene programs per density; the target-switch pairs, the cued levels and the probe on the same scenes are reported separately (Table~\ref{tab:hierarchy}). Pair accuracy is used for the grounding suite of Appendix~\ref{app:grounding}.

\textbf{Training selection and splits.} From each of 720 scene programs with 8 or 12 points the corpus takes four prompts, the L3 question and the probe on both twins (2,880 prompts, 960 per role). L1 and L2 prompts are never used in training, 20-point scenes are reserved for evaluation, and training, development and confirmatory scenes come from disjoint scene programs. The seed replication of Table~\ref{tab:seeds} adds an untouched evaluation set, described there. On the corpus, 133 of 46,080 image-free audit responses were correct; this is the blind reward opportunity of 0.003 reported in \S\ref{sec:measure}.

\textbf{Criteria and their outcome.} The confirmatory test of \S\ref{sec:verify} qualifies an evaluation cell (one scene density) by five criteria, listed with their outcome for every density in Table~\ref{tab:criteria}. Caption-only accuracy (criterion 4) uses question-blind captions written by Qwen2.5-VL-72B-Instruct (served with vLLM in bf16, temperature 0, at most 384 tokens); the untrained 7B model then answers from the caption alone. The artifact screen (criterion 5) trains four ridge-regularized linear classifiers ($\lambda=10^{-2}$) to tell the two twins of an L3 pair apart, from a DINOv2-small image embedding, the PNG file size, 21 image and frequency statistics, or 13 file features, on the target-stable and target-switch pairs (200 images per density) with five-fold cross-validation grouped by pair. A cell passes when, for every classifier, the out-of-fold $\max(\mathrm{AUC}, 1-\mathrm{AUC})$ is at most 0.55 and the upper end of the 95\% pair-bootstrap interval of the unfolded AUC is at most 0.62, at each of three cross-validation seeds, which change folds and bootstrap draws on the same images; passing the screen shows that these classifiers cannot tell the twins apart, not that no artifact exists.

The development set, used to track learning, meets all five criteria at no density, so acquisition is assessed on a separate confirmatory set of 450 new scene programs, 150 per density; its 20-point cell meets all five and supplies the confirmatory result (Table~\ref{tab:confirmatory}). The criteria and their numerical limits were set before the confirmatory set was generated; the requirement that the artifact screen pass at each of the three seeds was added after generation began and before any criterion was evaluated.

\textbf{Ordinary training.} Tables~\ref{tab:ladderdev} and~\ref{tab:tasklayer} give discovery and the task-layer gains of the models trained with ordinary rewards (\S\ref{sec:ladder}).
\begin{table}[h]
\caption{The three levels and the identification probe on one development scene (eight labeled points; the K8 scene of Figure~\ref{fig:design}(b)): the question templates, the cue each level supplies, and the gold answers on the two twins, which differ in one fact (K8 moves from $(-1,-1)$ to $(6,-1)$). L2 and the probe use the L3 image; L1 adds only the offset arrow.}
\label{tab:ladder}
\centering
\scriptsize
\setlength{\tabcolsep}{3pt}
\begin{tabular}{>{\raggedright\arraybackslash}p{2.9cm} >{\raggedright\arraybackslash}p{5.6cm} >{\raggedright\arraybackslash}p{1.7cm} c c}
\toprule
level & question & visual cue & gold (twin a) & gold (twin b) \\
\midrule
L1 (arrow cue; position given) & Point K8 has the smallest $y$-coordinate. What is the $x$-coordinate of point K8? & offset arrow at K8 & $-1$ & $6$ \\
L2 (target named; identity given) & Point K8 has the smallest $y$-coordinate. What is the $x$-coordinate of point K8? & none & $-1$ & $6$ \\
L3 (discover $+$ ground $+$ read; nothing given) & Consider the point with the smallest $y$-coordinate. What is its $x$-coordinate? & none & $-1$ & $6$ \\
probe (identification only) & Which labeled point has the smallest $y$-coordinate? & none & K8 & K8 \\
\bottomrule
\end{tabular}
\par\vspace{2pt}
{\scriptsize\raggedright Decoding is greedy with at most 32 new tokens. Results on these tasks are reported as member accuracy; pair accuracy (Eq.~(\ref{eq:pairacc})) is used for the grounding suite.\par}
\end{table}

\begin{table}[h]
\caption{Discovery on the coordinate development set (member accuracy with real images) for the models trained with ordinary rewards: the ViRL39K-trained 3B conditions (two seeds; step 100) and the Geometry3K-trained 7B pair (step 100), with the untrained models. Discovery (L3) and the identification probe at 8, 12, and 20 points. Two-seed means are computed from the per-seed counts; intervals are not reported for these values.}
\label{tab:ladderdev}
\centering
\scriptsize
\setlength{\tabcolsep}{3pt}
\begin{tabular}{l lll lll}
\toprule
model & discovery 8-pt & 12-pt & 20-pt & probe 8-pt & 12-pt & 20-pt \\
\midrule
\multicolumn{7}{l}{\textit{3B, ViRL39K-trained (step 100); two seeds}} \\
untrained 3B & 0.330 & 0.260 & 0.245 & 0.705 & 0.680 & 0.630 \\
\cond{Real}, two-seed mean & 0.383 & 0.338 & 0.330 & 0.728 & 0.700 & 0.610 \\
\quad seeds 1 / 2 & 0.380 / 0.385 & 0.330 / 0.345 & 0.330 / 0.330 & 0.725 / 0.730 & 0.710 / 0.690 & 0.620 / 0.600 \\
\cond{Caption}, two-seed mean & 0.370 & 0.323 & 0.340 & 0.728 & 0.715 & 0.635 \\
\quad seeds 1 / 2 & 0.365 / 0.375 & 0.325 / 0.320 & 0.345 / 0.335 & 0.715 / 0.740 & 0.710 / 0.720 & 0.610 / 0.660 \\
\cond{None}, two-seed mean & 0.368 & 0.303 & 0.318 & 0.730 & 0.700 & 0.638 \\
\quad seeds 1 / 2 & 0.365 / 0.370 & 0.300 / 0.305 & 0.305 / 0.330 & 0.740 / 0.720 & 0.705 / 0.695 & 0.650 / 0.625 \\
\cond{Gray}, two-seed mean & 0.373 & 0.298 & 0.315 & 0.713 & 0.683 & 0.625 \\
\quad seeds 1 / 2 & 0.370 / 0.375 & 0.295 / 0.300 & 0.320 / 0.310 & 0.725 / 0.700 & 0.685 / 0.680 & 0.630 / 0.620 \\
\addlinespace[2pt]\multicolumn{7}{l}{\textit{7B, Geometry3K-trained (step 100)}} \\
untrained 7B & 0.660 & 0.575 & 0.470 & 0.940 & 0.910 & 0.840 \\
\cond{Real} & 0.765 & 0.575 & 0.460 & 0.940 & 0.905 & 0.855 \\
\cond{Gray} & 0.720 & 0.570 & 0.460 & 0.945 & 0.905 & 0.845 \\
\bottomrule
\end{tabular}
\end{table}

\begin{table}[h]
\caption{Task-layer gains of the Geometry3K-trained 3B conditions at step 100 (pair accuracy; three-seed means with 95\% paired-item intervals; per-seed gains beneath). The equivalence test for the grounding task is in Appendix~\ref{app:stats}.}
\label{tab:tasklayer}
\centering
\scriptsize
\setlength{\tabcolsep}{3pt}
\begin{tabular}{l l l l}
\toprule
condition & cued readout: gain [95\% CI] & find-and-bind & header-cued table \\
\midrule
untrained 3B (pair accuracy) & 0.320 & 0.455 & 0.867 \\
\cond{Real} & $+$0.143 [0.102, 0.188] & $+$0.018 [$-$0.007, 0.044] & $+$0.019 [$-$0.002, 0.042] \\
\quad seeds 1 / 2 / 3 & $+$0.157 / $+$0.127 / $+$0.147 & $+$0.012 / $+$0.022 / $+$0.022 & $+$0.030 / $+$0.013 / $+$0.013 \\
\cond{Caption} & $+$0.107 [0.066, 0.149] & $+$0.006 [$-$0.018, 0.030] & $+$0.021 [$-$0.001, 0.044] \\
\quad seeds 1 / 2 / 3 & $+$0.130 / $+$0.113 / $+$0.077 & $+$0.003 / 0.000 / $+$0.013 & $+$0.023 / $+$0.027 / $+$0.013 \\
\cond{None} & $+$0.108 [0.070, 0.148] & $-$0.013 [$-$0.039, 0.013] & $+$0.023 [0.000, 0.049] \\
\quad seeds 1 / 2 / 3 & $+$0.093 / $+$0.097 / $+$0.133 & $-$0.010 / $-$0.010 / $-$0.020 & $+$0.023 / $+$0.020 / $+$0.027 \\
\cond{Gray} & $+$0.138 [0.096, 0.182] & $-$0.028 [$-$0.055, $-$0.002] & $+$0.023 [0.002, 0.047] \\
\quad seeds 1 / 2 / 3 & $+$0.140 / $+$0.137 / $+$0.137 & $-$0.030 / $-$0.032 / $-$0.023 & $+$0.027 / $+$0.017 / $+$0.027 \\
\bottomrule
\end{tabular}
\par\vspace{2pt}
{\scriptsize\raggedright Cued readout accounts for 72.1\% of \cond{Real}'s pair-count-weighted improvement across the 1,200-pair suite: its gain times its pair count, divided by the same product summed over the three tasks. Hierarchy discovery, measured by member accuracy, is not part of this denominator.\par}
\end{table}

\begin{table}[h]
\caption{Outcome of the five evaluation criteria of \S\ref{sec:verify} on the untrained 7B model, for every density of the development set (used to track learning) and of the separate confirmatory set: (1) discovery accuracy between 0.20 and 0.80; (2) discovery at least 0.05 below both cued levels; (3) exactly zero accuracy on discovery and its probe with a gray canvas and with no image; (4) caption-only accuracy at most half of with-image discovery accuracy; (5) the artifact screen of Appendix~\ref{app:ladder}: out-of-fold $\max(\mathrm{AUC},1-\mathrm{AUC})$ of four classifiers at most 0.55, and the upper end of the unfolded interval at most 0.62, at each of three cross-validation seeds. Criteria 1--3 are met in every cell. Only the 20-point confirmatory cell meets all five; it supplies the confirmatory result.}
\label{tab:criteria}
\centering
\scriptsize
\setlength{\tabcolsep}{3pt}
\begin{tabular}{l c r r r l l c}
\toprule
 &  &  & \multicolumn{3}{c}{(4) caption-only accuracy} &  & \\
\cmidrule(lr){4-6}
set & density & discovery (L3) & accuracy & limit ($\tfrac12$ L3) & outcome & (5) artifact screen & criteria met \\
\midrule
development & 8-pt & 0.660 & 0.580 & 0.330 & not met & met & 4 of 5 \\
 & 12-pt & 0.575 & 0.445 & 0.288 & not met & not met: 0.557 in 1 of 3 seeds & 3 of 5 \\
 & 20-pt & 0.470 & 0.185 & 0.235 & met & not met: 0.565 in 1 of 3 seeds & 4 of 5 \\
confirmatory & 8-pt & 0.585 & 0.570 & 0.293 & not met & met & 4 of 5 \\
 & 12-pt & 0.560 & 0.410 & 0.280 & not met & met & 4 of 5 \\
 & 20-pt & 0.425 & 0.130 & 0.213 & met & met & 5 of 5 \\
\bottomrule
\end{tabular}
\end{table}

\AppendixBarrier
\section{Independently Constructed Grounding Evaluations}\label{app:grounding}
Each task pairs two renders of one scene that differ in a single answer-relevant fact, so an answer that ignores the image cannot solve the pair (Proposition~\ref{prop:pairzero}). The tasks were built before the constructed training corpus existed and share no scenes with it. Table~\ref{tab:grounding} lists them with their roles and untrained levels and summarizes their validation.
\begin{table}[h]
\caption{The independently constructed grounding tasks: the 1,200-pair counterfactual suite (three tasks with distinct roles, reported separately) and its independently regenerated twin (fresh seeds, no iteration). Untrained pair accuracy (both members correct) for the 3B and 7B models; 7B levels with 95\% intervals. Validation: Qwen2.5-VL-72B-Instruct given question-blind captions reaches at most 0.054; every image-removed condition scores exactly 0.000, with all responses collapsing to one answer; 60 of 60 audited pairs were confirmed by human review; Gemma-3 and InternVL3-9B also score exactly 0.000 without the image.}
\label{tab:grounding}
\centering
\scriptsize
\setlength{\tabcolsep}{3pt}
\begin{tabular}{>{\raggedright\arraybackslash}p{3.6cm} r >{\raggedright\arraybackslash}p{3.3cm} r l}
\toprule
task & pairs & role & untrained 3B & untrained 7B [95\% CI] \\
\midrule
grounding task (named-point coordinate pairs) & 600 & find-and-bind (primary) & 0.455 & 0.768 [0.733, 0.802] \\
header-cued table task & 300 & cued control, high baseline & 0.867 & 0.993 [0.983, 1.000] \\
cued-readout task (marked-plot pairs) & 300 & cued readout, location marked & 0.320 & 0.673 [0.620, 0.727] \\
twin of the grounding task & 600 & regenerated twin & -- & 0.728 [0.692, 0.762] \\
twin of the cued-readout task & 300 & twin control & -- & 0.623 [0.570, 0.677] \\
twin of the header-cued table task & 300 & twin control & -- & 0.997 [0.990, 1.000] \\
\bottomrule
\end{tabular}
\par\vspace{2pt}
{\scriptsize\raggedright The twin was evaluated for the 7B models only.\par}
\end{table}

\AppendixBarrier
\section{Training and Canonical Access Results}\label{app:training}
\textbf{Recipes.} Table~\ref{tab:recipes} lists the training recipe of every run behind the main-text results. Training uses EasyR1 (revision dd71bbd) with local patches for the visual-access conditions; the repository lists the patches and the recorded revision of every run. The vision encoder and its projector into the language model are frozen in every run except the long-horizon runs, so only the language model is trained; the saved visual weights of the constructed-corpus checkpoints are identical to the base model's.

\textbf{Segmented training.} The constructed-corpus runs of \S\S\ref{sec:flip}--\ref{sec:dial} share an initial 30-step budget; the two \cond{Real} runs of seed 0 and the \cond{Real} run of seed 1 continue to 100 steps in segments: at steps 30, 50 and 75 a fresh trainer resumes from the previous checkpoint, so the AdamW moments reset, the KL reference is re-anchored to the resumed model, and the corpus order restarts from its beginning. Every matched 30-step comparison ends before the first restart.

\textbf{Reward received by the control.} The \cond{Gray}-trained control receives format reward throughout and only sparse correctness reward: its mean accuracy reward rises from 0.0008 at step 1 to 0.031 at step 30 (zero on probe rows), while its format reward rises from 0.87 to 1.00. In the three further seeds (Table~\ref{tab:seedrewards}), the control's mean accuracy reward is 0.0017--0.0050 at step 1 and 0.021--0.048 at step 30, against 0.972--0.991 for the \cond{Real} and named-target runs.

\textbf{Replication structure.} An independent seed changes two things here: the shuffled order in which the corpus is presented, and the seed of rollout sampling. The Geometry3K, ViRL39K and long-horizon runs draw both afresh for each seed. The two 100-step \cond{Real} constructed-corpus runs of seed 0 instead repeat one configuration, including its seeds and data order, so the pair measures how closely a repeated run reproduces the first; the corpus is presented in the same fixed order in both. Seeds 1--3 of the constructed-corpus conditions (\cond{Real}, \cond{Gray}-trained, named-target and shortcut-accessible; Tables~\ref{tab:seeds} and~\ref{tab:shortcut}) draw both afresh, with data-order and rollout-sampling seeds 1/1, 2/2 and 3/3 against 0/0 for seed 0, and run 30 steps each; the \cond{Real} run of seed 1 then continues to step 100 on the seed-0 schedule. Table~\ref{tab:recipes} therefore counts seeds and runs in separate columns, and Table~\ref{tab:runbasis} gives the seeds, the runs and the resampled unit behind each result reported in this paper.

Table~\ref{tab:table1perseed} gives Table~\ref{tab:access} per seed; the \cond{Caption} condition and the caption test input use question-blind captions written by Qwen2.5-VL-3B-Instruct. Table~\ref{tab:geo7b} gives the counts behind the 7B share of \S\ref{sec:blind}. Table~\ref{tab:access7b} gives the 7B pair on the independently constructed grounding suite and its twin (\S\ref{sec:scale}). Table~\ref{tab:extract} splits the blind gains of Table~\ref{tab:access} by whether the untrained model's real-image answer could be extracted (497 of 601 questions); its note defines extractability.
\begin{table}[h]
\caption{Training recipes. Every run uses GRPO with 5 rollouts per prompt, learning rate $10^{-6}$ and KL coefficient 0.01. In every recipe the verifier scores each response's extracted answer against its gold answer and, where noted, its format; no image-dependent auxiliary loss or pairwise training reward is added. Constructed-corpus runs (\S\S\ref{sec:flip}--\ref{sec:dial}): rollout batch 240, global batch 48. Geometry3K, ViRL39K and long-horizon runs: rollout batch 512, global batch 128 (the long-horizon runs on 4 GPUs with tensor parallelism 2). The last two columns count, per condition, the independent seeds and the training runs: an independent seed draws its own shuffled data order and rollout sampling; a run beyond the seed count repeats one configuration. The dose response of \S\ref{sec:dose} has four points, the three mixtures listed here and the standard run at step 30 ($f=1$).}
\label{tab:recipes}
\centering
\scriptsize
\setlength{\tabcolsep}{3pt}
\begin{tabular}{>{\raggedright\arraybackslash}p{2.1cm} >{\raggedright\arraybackslash}p{1.0cm} >{\raggedright\arraybackslash}p{1.6cm} >{\raggedright\arraybackslash}p{2.2cm} >{\raggedright\arraybackslash}p{1.9cm} >{\raggedright\arraybackslash}p{1.7cm} >{\raggedright\arraybackslash}p{0.6cm} >{\raggedright\arraybackslash}p{0.9cm}}
\toprule
experiment & backbone, vision encoder & training images & reward & training data & steps; evaluated at & seeds & runs \\
\midrule
Geometry3K access conditions (\S\ref{sec:blind}, \S\ref{sec:same-items}) & 3B, frozen & \cond{Real} / \cond{Gray} / \cond{None} / \cond{Caption}, one condition per run & answer correctness & Geometry3K, filtered (1,288 items) & 100 & 3 & 3 \\
7B access pair (\S\ref{sec:blind}, \S\ref{sec:scale}) & 7B, frozen & \cond{Real}; \cond{Gray} & answer correctness & Geometry3K & 100 & 1 & 1 \\
long-horizon runs (\S\ref{sec:decay}) & 3B, trainable & real images & $0.5\cdot$accuracy $+\;0.5\cdot$format & Geometry3K, unfiltered & 400, in segments; evaluated at 100, 150, 200, 300, 400 & 2 & 2 \\
resolvable training, standard reward (\S\ref{sec:flip}) & 7B, frozen & real images & $0.5\cdot$accuracy $+\;0.5\cdot$format & constructed corpus, 2,880 items (discovery and probe, both twins) & seeds 0 and 1: 100, in segments (seed 0 evaluated at 10, 20, 30, 50, 75, 100; seed 1 at 30 and 100). Seeds 2 and 3: 30 & 4 & 5 (two of seed 0) \\
\cond{Gray}-trained control (\S\ref{sec:drives}) & 7B, frozen & gray canvases & as above & same corpus & 30; evaluated at 10, 20, 30 (seed 0) or 30 (seeds 1--3) & 4 & 4 \\
named-target ablation (\S\ref{sec:acquired}; Table~\ref{tab:seeds}) & 7B, frozen & real images & as above & same scene programs, with the find-and-bind (L2) question in place of the discovery question; probe unchanged & 30 & 3 & 3 \\
shortcut-accessible twin (\S\ref{sec:drives}; Table~\ref{tab:shortcut}) & 7B, frozen & real images, with the question-blind 72B caption of each training image in the prompt & as above & same 2,880 prompts; caption placed after the image and before the question & 30 & 3 & 3 \\
3B capacity comparison (\S\ref{sec:capacity}) & 3B, frozen & real images & as the standard reward & constructed corpus, identical scenes & 30; evaluated at 10, 20, 30 & 1 & 1 \\
mixtures $f = 0, \tfrac13, \tfrac23$ (\S\ref{sec:dose}) & 7B, frozen & real images & as the standard reward & constructed corpus mixed with decontaminated ViRL39K in 240-prompt steps & 30; evaluated at 10, 20, 30 & 1 & 1 each \\
ViRL39K access conditions (\S\ref{sec:ladder}) & 3B, frozen & \cond{Real} / \cond{Gray} / \cond{None} / \cond{Caption}, one condition per run & answer correctness & decontaminated ViRL39K, single-image items (23,542) & 100 & 2 & 2 \\
\bottomrule
\end{tabular}
\par\vspace{2pt}
{\scriptsize\raggedright Segments. In the 100-step constructed-corpus runs, each restart (steps 30, 50 and 75) warm-starts a fresh trainer from the merged checkpoint, with the effects described above; the step-30 checkpoint, re-evaluated after the extension, matched its original evaluation, and every 30-step run (\cond{Gray}-trained control, 3B comparison, mixtures) ends before the first restart. The long-horizon runs were also trained in segments (seed 1 continues a 100-step run to step 400 in 50-step segments; seed 2 runs from the base model to step 400).\par}
\end{table}

\begin{table}[h]
\caption{Table~\ref{tab:access} per seed, with the untrained and trained models scored by the same answer matcher: accuracy of each condition's step-100 model under the four test conditions (601 Geometry3K test items), the gain over the untrained model under the same test condition, and the recovery share (real-tested gain $\div$ \cond{Real}'s real-tested gain). Intervals are not computed for the three-seed means and shares; the per-seed values show their spread.}
\label{tab:table1perseed}
\centering
\scriptsize
\setlength{\tabcolsep}{3pt}
\begin{tabular}{l c rrrr rrrr r}
\toprule
condition & seed & acc.\ real & acc.\ gray & acc.\ none & acc.\ caption & gain real & gain gray & gain none & gain caption & share \\
\midrule
\textit{base model} & -- & 0.175 & 0.090 & 0.068 & 0.210 & -- & -- & -- & -- & -- \\
\cond{Real} & 1 & 0.423 & 0.106 & 0.108 & 0.311 & $+$0.248 & $+$0.017 & $+$0.040 & $+$0.102 & -- \\
 & 2 & 0.419 & 0.112 & 0.098 & 0.301 & $+$0.245 & $+$0.022 & $+$0.030 & $+$0.091 & -- \\
 & 3 & 0.398 & 0.110 & 0.106 & 0.326 & $+$0.223 & $+$0.020 & $+$0.038 & $+$0.117 & -- \\
\cond{Caption} & 1 & 0.361 & 0.118 & 0.112 & 0.318 & $+$0.186 & $+$0.028 & $+$0.043 & $+$0.108 & 0.752 \\
 & 2 & 0.346 & 0.108 & 0.106 & 0.291 & $+$0.171 & $+$0.018 & $+$0.038 & $+$0.082 & 0.701 \\
 & 3 & 0.331 & 0.117 & 0.097 & 0.326 & $+$0.156 & $+$0.027 & $+$0.028 & $+$0.117 & 0.702 \\
\cond{None} & 1 & 0.325 & 0.095 & 0.097 & 0.286 & $+$0.150 & $+$0.005 & $+$0.028 & $+$0.076 & 0.604 \\
 & 2 & 0.308 & 0.117 & 0.121 & 0.294 & $+$0.133 & $+$0.027 & $+$0.053 & $+$0.085 & 0.544 \\
 & 3 & 0.303 & 0.103 & 0.118 & 0.261 & $+$0.128 & $+$0.013 & $+$0.050 & $+$0.052 & 0.575 \\
\cond{Gray} & 1 & 0.311 & 0.110 & 0.108 & 0.286 & $+$0.136 & $+$0.020 & $+$0.040 & $+$0.076 & 0.550 \\
 & 2 & 0.306 & 0.097 & 0.088 & 0.270 & $+$0.131 & $+$0.007 & $+$0.020 & $+$0.060 & 0.537 \\
 & 3 & 0.281 & 0.106 & 0.103 & 0.245 & $+$0.106 & $+$0.017 & $+$0.035 & $+$0.035 & 0.478 \\
\bottomrule
\end{tabular}
\end{table}

\begin{table}[h]
\caption{The ordinary 7B access pair of \S\ref{sec:scale} on the independently constructed grounding suite and its twin (Geometry3K training, step 100): untrained pair accuracy and the change under \cond{Real} and \cond{Gray} training, with 95\% paired-item bootstrap intervals (2,000 draws). The interval excludes zero for \cond{Real} on the grounding task and on its twin, and includes zero in every other cell.}
\label{tab:access7b}
\centering
\scriptsize
\setlength{\tabcolsep}{3pt}
\begin{tabular}{l r r l l}
\toprule
task & pairs & untrained 7B & \cond{Real} $\Delta$ [95\% CI] & \cond{Gray} $\Delta$ [95\% CI] \\
\midrule
grounding task & 600 & 0.768 & $+$0.035 [0.013, 0.058] & $+$0.012 [$-$0.010, 0.033] \\
twin of the grounding task & 600 & 0.728 & $+$0.038 [0.017, 0.062] & $+$0.013 [$-$0.010, 0.037] \\
cued-readout task & 300 & 0.673 & $+$0.030 [$-$0.007, 0.067] & $+$0.017 [$-$0.013, 0.047] \\
twin of the cued-readout task & 300 & 0.623 & $+$0.013 [$-$0.030, 0.060] & $+$0.013 [$-$0.023, 0.050] \\
header-cued table task & 300 & 0.993 & 0.000 [$-$0.010, 0.010] & $-$0.003 [$-$0.010, 0.000] \\
twin of the header-cued table task & 300 & 0.997 & 0.000 [$-$0.010, 0.010] & $-$0.007 [$-$0.017, 0.000] \\
\bottomrule
\end{tabular}
\end{table}

\begin{table}[h]
\caption{The 7B access pair behind the share in Table~\ref{tab:access} (Geometry3K training; 601 Geometry3K test items; the answer matcher of Table~\ref{tab:access}): correct answers with real test images and with a gray canvas, with 95\% item-bootstrap intervals, and the gain over the untrained model with real images, with its item-paired interval (5,000 draws). The recovery share is \cond{Gray}'s gain divided by \cond{Real}'s: 116/149 = 0.779 [0.642, 0.921].}
\label{tab:geo7b}
\centering
\scriptsize
\setlength{\tabcolsep}{3pt}
\begin{tabular}{l c >{\raggedright\arraybackslash}p{3.2cm} >{\raggedright\arraybackslash}p{3.0cm} >{\raggedright\arraybackslash}p{3.2cm}}
\toprule
trained with & step & real image: correct [95\% CI] & gain over untrained [95\% CI] & gray canvas: correct [95\% CI] \\
\midrule
untrained 7B & -- & 141/601 = 0.235 [0.201, 0.270] & -- & 48/601 = 0.080 [0.058, 0.103] \\
\cond{Real} & 100 & 290/601 = 0.483 [0.443, 0.521] & $+$0.248 [0.203, 0.291] (149/601) & 75/601 = 0.125 [0.100, 0.151] \\
\cond{Gray} & 100 & 257/601 = 0.428 [0.388, 0.466] & $+$0.193 [0.153, 0.235] (116/601) & 79/601 = 0.131 [0.103, 0.160] \\
\bottomrule
\end{tabular}
\end{table}

\begin{table}[h]
\caption{The blind gains of Table~\ref{tab:access} by whether the untrained 3B model's real-image answer can be extracted (extractable: 497 of 601 questions): the untrained model's accuracy, the gain over it with real test images, averaged over the three seeds per question, and each condition's share of \cond{Real}'s gain, with 95\% intervals from a 10,000-draw bootstrap over questions. The three runs of a condition answer the same 601 questions, so questions, not the 1,803 responses, are resampled.}
\label{tab:extract}
\centering
\scriptsize
\setlength{\tabcolsep}{3pt}
\begin{tabular}{l r r l l l l}
\toprule
questions & $n$ & untrained & \cond{Real} gain & \cond{Caption} gain & \cond{None} gain & \cond{Gray} gain \\
\midrule
all questions & 601 & 0.175 & $+$0.238 [0.202, 0.275] & $+$0.171 [0.136, 0.207] & $+$0.137 [0.103, 0.171] & $+$0.125 [0.092, 0.158] \\
extractable & 497 & 0.211 & $+$0.200 [0.160, 0.240] & $+$0.140 [0.100, 0.179] & $+$0.107 [0.069, 0.145] & $+$0.101 [0.064, 0.139] \\
not extractable & 104 & 0.000 & $+$0.423 [0.343, 0.503] & $+$0.324 [0.250, 0.401] & $+$0.279 [0.212, 0.349] & $+$0.237 [0.176, 0.301] \\
\bottomrule
\end{tabular}
\par\medskip
\begin{tabular}{l l l l}
\toprule
questions & \cond{Caption} share & \cond{None} share & \cond{Gray} share \\
\midrule
all questions & 0.719 [0.614, 0.821] & 0.574 [0.467, 0.673] & 0.523 [0.416, 0.622] \\
extractable & 0.698 [0.565, 0.828] & 0.537 [0.395, 0.664] & 0.507 [0.364, 0.636] \\
not extractable & 0.765 [0.620, 0.930] & 0.659 [0.519, 0.813] & 0.561 [0.438, 0.691] \\
\bottomrule
\end{tabular}
\par\vspace{2pt}
{\scriptsize\raggedright An answer is extractable when the answer extractor returns one, which is distinct from following the answer-tag format. The untrained model's answers are extractable for 497 questions with the image and 502 with the caption, with accuracy 0.211 and 0.251 among them, so its higher caption accuracy is not a difference in extraction coverage. The split is descriptive: it shows that blind gains are not confined to questions whose answer the untrained model could not state, not how much of any gain is due to format.\par}
\end{table}

\begin{table}[h]
\caption{Run basis of each reported result. \emph{Seeds} counts the independent seeds of a condition and \emph{runs} its training runs (Table~\ref{tab:recipes}); where runs exceed seeds, the further run repeats one configuration. The fourth column names the table that prints the individual seeds or runs, and the last the unit that each result's 95\% intervals resample with the trained models held fixed.}
\label{tab:runbasis}
\centering
\scriptsize
\setlength{\tabcolsep}{3pt}
\begin{tabular}{>{\raggedright\arraybackslash}p{5.4cm} r >{\raggedright\arraybackslash}p{1.5cm} >{\raggedright\arraybackslash}p{2.8cm} >{\raggedright\arraybackslash}p{2.3cm}}
\toprule
result (section; table) & seeds & runs & individual seeds or runs & 95\% intervals resample \\
\midrule
blind recovery shares, 3B (\S\ref{sec:blind}; Table~\ref{tab:access}) & 3 & 3 & Table~\ref{tab:table1perseed}, per seed & point estimates over seeds \\
\cond{Gray} recovery share, 7B (\S\ref{sec:blind}; Table~\ref{tab:geo7b}) & 1 & 1 & -- & benchmark items \\
operation-level gains, 3B (\S\ref{sec:ladder}; Table~\ref{tab:tasklayer}) & 3 & 3 & Table~\ref{tab:tasklayer}, per seed & grounding pairs \\
discovery, ViRL39K-trained 3B (\S\ref{sec:ladder}; Table~\ref{tab:ladderdev}) & 2 & 2 & Table~\ref{tab:ladderdev}, per seed & point estimates over seeds \\
grounding loss under prolonged training (\S\ref{sec:decay}; Table~\ref{tab:trajectories}) & 2 & 2 & Table~\ref{tab:trajectories}, per run & grounding pairs \\
discovery on the confirmatory set (\S\ref{sec:acquired}; Tables~\ref{tab:confirmatory}, \ref{tab:seeds}) & 4 & 5 & Table~\ref{tab:confirmatory} (seed 0, per run); Table~\ref{tab:seeds} (seeds 1--3; seed 1 also at step 100) & scene programs (seed 0); seeds, $t$-based (Table~\ref{tab:seedcontrasts}) \\
visually attributable gain $\tau_{\mathrm{vis}}$ (\S\ref{sec:drives}; Tables~\ref{tab:attribution}, \ref{tab:seedcontrasts}) & 4 & 5 (\cond{Real}), 4 (\cond{Gray}) & Table~\ref{tab:seeds}, per seed & scene programs (seed 0); seeds, $t$-based (Table~\ref{tab:seedcontrasts}) \\
transfer to the grounding tasks (\S\ref{sec:transfer}; Tables~\ref{tab:transfer}, \ref{tab:seedcontrasts}) & 4 & 5 & Table~\ref{tab:transfer} (seed 0, per run); Table~\ref{tab:seeds} (seeds 1--3; seed 1 also at step 100) & grounding pairs (seed 0); seeds (Table~\ref{tab:seedcontrasts}) \\
named-target ablation (\S\ref{sec:acquired}; Table~\ref{tab:seedcontrasts}) & 3 & 3 per condition & Table~\ref{tab:seeds}, per seed & seeds, $t$-based (Table~\ref{tab:seedcontrasts}) \\
shortcut-accessible twin (\S\ref{sec:drives}; Tables~\ref{tab:shortcut}, \ref{tab:shortcutcontrasts}) & 3 & 3 & Table~\ref{tab:shortcut}, per seed & seeds, $t$-based (Table~\ref{tab:shortcutcontrasts}) \\
mixture contrasts at step 30 (\S\ref{sec:dose}; Tables~\ref{tab:dose} and~\ref{tab:slopes}) & 1 & 1 per mixture & -- & scene programs \\
3B against 7B (\S\ref{sec:capacity}; Table~\ref{tab:capacity}) & 1 & 1 per model & -- & scene programs \\
\bottomrule
\end{tabular}
\end{table}

\begin{table}[h]
\caption{Training rewards of seeds 1--3 (Table~\ref{tab:seeds}): the overall reward of Eq.~(\ref{eq:reward}) with its format and accuracy components at steps 1 and 30, and the mean accuracy component over steps 19--30. The \cond{Gray}-trained control earns format reward from the start and almost no accuracy reward at any point; the \cond{Real}, named-target and shortcut-accessible runs approach the ceiling of both components by step 30.}
\label{tab:seedrewards}
\centering
\scriptsize
\setlength{\tabcolsep}{3pt}
\begin{tabular}{l c l l r}
\toprule
condition & seed & step 1: overall / format / accuracy & step 30: overall / format / accuracy & mean accuracy, steps 19--30 \\
\midrule
\cond{Real} & 1 & 0.5975 / 0.8658 / 0.3292 & 0.9858 / 1.0000 / 0.9717 & 0.975 \\
\cond{Real} & 2 & 0.5900 / 0.8875 / 0.2925 & 0.9863 / 1.0000 / 0.9725 & 0.982 \\
\cond{Real} & 3 & 0.6095 / 0.8825 / 0.3365 & 0.9930 / 0.9992 / 0.9867 & 0.970 \\
\cond{Gray}-trained & 1 & 0.4350 / 0.8683 / 0.0017 & 0.5175 / 1.0000 / 0.0350 & 0.033 \\
\cond{Gray}-trained & 2 & 0.4413 / 0.8800 / 0.0025 & 0.5238 / 0.9992 / 0.0483 & 0.041 \\
\cond{Gray}-trained & 3 & 0.4425 / 0.8800 / 0.0050 & 0.5104 / 1.0000 / 0.0208 & 0.024 \\
named-target & 1 & 0.5829 / 0.8708 / 0.2950 & 0.9942 / 1.0000 / 0.9883 & 0.987 \\
named-target & 2 & 0.5810 / 0.8825 / 0.2795 & 0.9896 / 0.9992 / 0.9800 & 0.979 \\
named-target & 3 & 0.5998 / 0.8855 / 0.3140 & 0.9954 / 1.0000 / 0.9908 & 0.991 \\
shortcut-accessible & 1 & 0.7138 / 0.8892 / 0.5383 & 0.9904 / 1.0000 / 0.9808 & 0.977 \\
shortcut-accessible & 2 & 0.7000 / 0.8825 / 0.5175 & 0.9850 / 0.9992 / 0.9708 & 0.972 \\
shortcut-accessible & 3 & 0.7354 / 0.9008 / 0.5700 & 0.9946 / 1.0000 / 0.9892 & 0.981 \\
\bottomrule
\end{tabular}
\par\vspace{2pt}
{\scriptsize\raggedright Seed 1, continued to step 100 (Table~\ref{tab:seeds}): accuracy component 0.9875, 0.9933 and 0.9908 at steps 50, 75 and 100, format 1.0000 throughout, KL 0.041--0.054.\par}
\end{table}

\section{Corrosion}\label{app:corrosion}
Table~\ref{tab:trajectories} gives both long-horizon trajectories of \S\ref{sec:decay}: benchmark accuracy peaks at step 200, and grounding pair accuracy falls from step 100 to step 400 by 0.070 (95\% CI [0.043, 0.098]) and 0.052 ([0.020, 0.083]), ending 0.048 ([0.017, 0.082]) and 0.042 ([0.005, 0.078]) below the base model. In the first run, 137 of 601 benchmark items change correctness between steps 100 and 400 despite a net gain of three correct answers: a flat score, a churning model. Relative to \S\ref{sec:blind}, these runs change the horizon, the encoder, the corpus filter and the reward implementation together, so they show the loss under this recipe rather than isolating one cause.

Table~\ref{tab:overlap} gives the per-run degraded sets, their overlap against the permutation null, and the shared wrong answers behind Figure~\ref{fig:necessity}(c); its note characterizes those answers.

\begin{table}[h]
\caption{The two long-horizon runs of \S\ref{sec:decay} (Qwen2.5-VL-3B, trainable encoder, real images, unfiltered Geometry3K): benchmark accuracy (correct answers out of 601) with 95\% item-bootstrap intervals, and grounding-task pair accuracy with its interval and its change from the untrained model (the 600 pairs, each its own scene, resampled with the same 10,000 draws at every checkpoint). Peaks are marked.}
\label{tab:trajectories}
\centering
\scriptsize
\setlength{\tabcolsep}{3pt}
\begin{tabular}{c c l l l}
\toprule
seed & step & benchmark [95\% CI] & grounding [95\% CI] & $\Delta$ vs base [95\% CI] \\
\midrule
1 & 0 & 0.175 [0.145, 0.206] & 0.455 [0.415, 0.495] & -- \\
 & 100 & 0.431 [0.393, 0.471] & 0.477 [0.437, 0.517] & $+$0.022 [$-$0.008, 0.052] \\
 & 150 & 0.463 [0.424, 0.504] & 0.465 [0.425, 0.505] & $+$0.010 [$-$0.020, 0.040] \\
 & 200 & 0.483 [0.444, 0.524] (peak) & 0.450 [0.412, 0.490] & $-$0.005 [$-$0.035, 0.025] \\
 & 300 & 0.464 [0.424, 0.506] & 0.440 [0.400, 0.480] & $-$0.015 [$-$0.047, 0.017] \\
 & 400 & 0.436 [0.398, 0.474] & 0.407 [0.367, 0.447] & $-$0.048 [$-$0.082, $-$0.017] \\
2 & 0 & 0.175 [0.145, 0.206] & 0.455 [0.415, 0.495] & -- \\
 & 100 & 0.428 [0.389, 0.468] & 0.465 [0.425, 0.505] & $+$0.010 [$-$0.020, 0.040] \\
 & 150 & 0.446 [0.404, 0.488] & 0.468 [0.428, 0.508] & $+$0.013 [$-$0.020, 0.047] \\
 & 200 & 0.496 [0.454, 0.537] (peak) & 0.447 [0.407, 0.487] & $-$0.008 [$-$0.042, 0.025] \\
 & 300 & 0.449 [0.409, 0.488] & 0.400 [0.362, 0.438] & $-$0.055 [$-$0.090, $-$0.018] \\
 & 400 & 0.453 [0.411, 0.493] & 0.413 [0.373, 0.453] & $-$0.042 [$-$0.078, $-$0.005] \\
\bottomrule
\end{tabular}
\par\vspace{2pt}
{\scriptsize\raggedright From step 100 to step 400, grounding-task pair accuracy changes by $-$0.070 [$-$0.098, $-$0.043] in the first run and $-$0.052 [$-$0.083, $-$0.020] in the second.\par}
\end{table}

\begin{table}[h]
\caption{Shared errors across the three \cond{Gray}-trained runs of \S\ref{sec:same-items} (grounding task, 600 pairs). The sets are defined with the earlier answer-matcher revision (Appendix~\ref{app:stats}). Top: per run, the degraded set (pairs the untrained model solves and the run fails), pairs gained, and wrong members (individual pair members answered incorrectly within the degraded set). Bottom: overlap of the degraded sets (Jaccard; intersection/union) against a 10,000-draw permutation null of sets of the same sizes ($p = 10^{-4}$ for every comparison), and the rate of identical wrong answers on the wrong members the runs share.}
\label{tab:overlap}
\centering
\scriptsize
\setlength{\tabcolsep}{3pt}
\begin{tabular}{l r r r}
\toprule
run & degraded pairs & pairs gained & wrong members \\
\midrule
seed 1 & 51 & 24 & 52 \\
seed 2 & 49 & 22 & 53 \\
seed 3 & 45 & 23 & 46 \\
\bottomrule
\end{tabular}
\par\medskip
\begin{tabular}{l l l l}
\toprule
sets & Jaccard & permutation null mean & identical wrong answer, shared wrong members [95\% CI] \\
\midrule
seeds 1 vs 2 & 0.724 (42/58) & 0.097 (95th pct.\ 0.149) & 0.976 [0.877, 0.996] (41/42) \\
seeds 1 vs 3 & 0.811 (43/53) & 0.093 (95th pct.\ 0.143) & 1.000 [0.920, 1.000] (44/44) \\
seeds 2 vs 3 & 0.741 (40/54) & 0.091 (95th pct.\ 0.146) & 0.976 [0.874, 0.996] (40/41) \\
all three & 0.661 (39/59) & 0.011 (95th pct.\ 0.026) & 0.975 [0.871, 0.996] (39/40) \\
\bottomrule
\end{tabular}
\par\vspace{2pt}
{\scriptsize\raggedright Degraded sets are drawn from the 283 of 600 pairs the untrained model solves under that matcher revision. Pairs failed by seed 1 / 2 / 3 only: 5 / 6 / 1; by exactly two seeds (1\&2 / 1\&3 / 2\&3): 3 / 4 / 1; by all three: 39; union 59. The 39 identical answers take 11 distinct values; the most frequent value, 1, occurs 14 times, each time against a gold answer of $-$1. Several explanations drawn from the scene can fit one answer, and each answer is counted under every explanation it fits: the gold plus one, skipping zero (21); the gold with its sign flipped (15); the nearest point's $x$-coordinate (6); the target's own $y$-coordinate (3). Eleven answers match only some coordinate of the scene, which on a dense scene does not show that the point was selected. On the same 39 member slots, the three \cond{Real}-trained runs give the identical wrong answer in 29 slots in at least one run and in 20 in all three (\cond{None}: 37 and 28; \cond{Caption}: 32 and 21). Re-scored with the current matcher, the degraded sets have 48, 47 and 43 pairs, 37 of them shared by all three runs.\par}
\end{table}

\section{Gains, Attribution, and Mixture Measurement}\label{app:measures}
Fix a task, scoring rule, model size, and checkpoint budget. Let $a_c^{v}(t)$ be the accuracy of training condition $c$ at step $t$ under evaluation input $v$, and $a_0^{v}$ that of the base model. The gain and the visually attributable gain of \S\ref{sec:drives} are
\begin{equation}
g_c(t)=a_c^{\mathrm{image}}(t)-a_0^{\mathrm{image}},
\qquad
\tau_{\mathrm{vis}}(t)=g_{\mathrm{REAL}}(t)-g_{\mathrm{GRAY}}(t)
=a_{\mathrm{REAL}}^{\mathrm{image}}(t)-a_{\mathrm{GRAY}}^{\mathrm{image}}(t).
\label{eq:gains}
\end{equation}
Both trained models are evaluated with real images at the matched step, so the shared base cancels. The interval on $\tau_{\mathrm{vis}}$ comes from a paired comparison of the two models on the same items, not from arithmetic on the endpoints of the two marginal intervals. Table~\ref{tab:seedcontrasts} reports the same contrast per seed, with a $t$-based interval over seeds (Appendix~\ref{app:stats}). Test-time dependence is a different contrast: it changes the evaluation input of one checkpoint, $a_c^{\mathrm{image}}(t)-a_c^{\mathrm{removed}}(t)$, with the removal named (test-image deletion in \S\ref{sec:gains}, a gray canvas in \S\S\ref{sec:flip}--\ref{sec:dial}). The recovery share of \S\ref{sec:blind} is $g_c/g_{\mathrm{REAL}}$ with gains averaged over seeds before division; it is a ratio of averaged gains, not an average of per-seed ratios, and it is not confined to $[0,1]$.

The measured resolvability $m(f)$ of Eq.~(\ref{eq:mass}), the audited mass $m(f)$ of the mixture in Appendix~\ref{app:dose}, combines two component means. With $\mathcal C$ the 2,880 audited constructed training rows and $\mathcal V_f$ the unique audited ViRL39K items in the stream of condition $f$,
\begin{equation}
\overline{\Delta q}_{\mathrm{constructed}}
=\frac1{|\mathcal C|}\sum_{i\in\mathcal C}\Delta q_i,
\qquad
\overline{\Delta q}_{\mathrm{ViRL},f}
=\frac1{|\mathcal V_f|}\sum_{i\in\mathcal V_f}\Delta q_i .
\label{eq:massmeans}
\end{equation}
The constructed mean is the same in every condition, and the ordinary mean depends on $f$; at $f=1$ the mass is the constructed mean, and no mean over an empty ordinary subset is evaluated. Both means weight unique audited items equally. The standardization of Appendix~\ref{app:standardization} uses a different audit population, the realized training rows weighted by their presentations.

\AppendixBarrier
\section{Acquisition, Attribution, and Transfer}\label{app:acquisition}
Table~\ref{tab:devgrid} gives development-set acquisition at steps 0, 30 and 100 for both runs, with the identification probe; Table~\ref{tab:confirmatory} every confirmatory-set cell; Table~\ref{tab:attribution} the attribution of \S\ref{sec:drives} on both sets; Table~\ref{tab:transfer} transfer to the independently constructed grounding tasks at the same three checkpoints; and Table~\ref{tab:hierarchy} the same-scene hierarchy and the target-switch pairs on the 20-point confirmatory scenes. The development trajectories at all six checkpoints are plotted in Figure~\ref{fig:flip}(a); Figure~\ref{fig:flip}(c) shows the transfer changes at step 100. Tables~\ref{tab:seeds} and~\ref{tab:seedcontrasts} give the seed replication of the matched comparison and the named-target ablation; Table~\ref{tab:composite} gives the composite-operation analysis, and Tables~\ref{tab:shortcut} and~\ref{tab:shortcutcontrasts} the shortcut-accessible twin.

A gray canvas does not make the models refuse; it makes them give the same default answer every time. With a gray canvas, the \cond{Real} runs and the untrained model answer every question in the required format but with a fixed default: 0 on discovery, which is not among the 14 possible coordinates, and a letter that labels no point on the probe. Every such cell therefore scores zero; the responses are not refusals, which occur only when the image is removed altogether. The seed-0 \cond{Gray}-trained control and the 3B model can default to a coordinate that occurs and score a few percent on discovery with a gray canvas (Tables~\ref{tab:confirmatory}, \ref{tab:attribution} and~\ref{tab:capacity}); on the 20-point confirmatory set the control and the \cond{Gray}-trained seeds 1--3 score 0 of 200.
\begin{table}[h]
\caption{Development-set acquisition (Figure~\ref{fig:flip}(a)): discovery (L3) member accuracy with real images at steps 0, 30 and 100 for both runs, by density, and its change from the untrained model, with 95\% intervals from resampling scene programs (Appendix~\ref{app:stats}); last column, the identification probe with real images at 8 / 12 / 20 points. With the test image replaced by a gray canvas, discovery and the probe score 0 of 200 at every one of the six checkpoints of both runs.}
\label{tab:devgrid}
\centering
\scriptsize
\setlength{\tabcolsep}{3pt}
\begin{tabular}{l c >{\raggedright\arraybackslash}p{2.9cm} >{\raggedright\arraybackslash}p{2.9cm} >{\raggedright\arraybackslash}p{2.9cm} l}
\toprule
run & step & 8-pt: level [95\% CI]; $\Delta$ [95\% CI] & 12-pt & 20-pt & probe 8 / 12 / 20-pt \\
\midrule
untrained 7B & 0 & 0.660 [0.580, 0.740] & 0.575 [0.485, 0.660] & 0.470 [0.385, 0.555] & 0.940 / 0.910 / 0.840 \\
run 1 & 30 & 0.965 [0.935, 0.990]\newline $\Delta$ $+$0.305 [0.230, 0.380] & 0.890 [0.830, 0.945]\newline $\Delta$ $+$0.315 [0.230, 0.400] & 0.800 [0.730, 0.865]\newline $\Delta$ $+$0.330 [0.255, 0.410] & 1.000 / 0.975 / 0.975 \\
 & 100 & 200/200 (100 programs)\newline $\Delta$ $+$0.340 [0.260, 0.420] & 0.940 [0.895, 0.980]\newline $\Delta$ $+$0.365 [0.280, 0.455] & 0.910 [0.860, 0.950]\newline $\Delta$ $+$0.440 [0.360, 0.525] & 1.000 / 0.990 / 0.995 \\
run 2 & 30 & 0.960 [0.925, 0.985]\newline $\Delta$ $+$0.300 [0.225, 0.380] & 0.860 [0.795, 0.920]\newline $\Delta$ $+$0.285 [0.205, 0.370] & 0.795 [0.730, 0.860]\newline $\Delta$ $+$0.325 [0.245, 0.405] & 1.000 / 0.960 / 0.960 \\
 & 100 & 0.995 [0.985, 1.000]\newline $\Delta$ $+$0.335 [0.255, 0.415] & 0.950 [0.915, 0.980]\newline $\Delta$ $+$0.375 [0.290, 0.465] & 0.905 [0.860, 0.945]\newline $\Delta$ $+$0.435 [0.350, 0.520] & 0.995 / 0.985 / 0.990 \\
\bottomrule
\end{tabular}
\end{table}

\begin{table}[h]
\caption{All confirmatory-set cells (150 new scene programs per density, generated after the evaluation criteria of \S\ref{sec:verify} were set): discovery member accuracy with real images and its change from the untrained model, and accuracy with the test image replaced by a gray canvas, with 95\% intervals from resampling scene programs; a cell with 0 of 200 correct is given as its count and number of scene programs. The last column gives how many of the five criteria the evaluation cell met before training; it describes the cell, not a run's outcome. The 20-point cell met all five and supplies the confirmatory result of \S\ref{sec:acquired}; the 8- and 12-point cells did not meet the caption criterion (Appendix~\ref{app:ladder}).}
\label{tab:confirmatory}
\centering
\scriptsize
\setlength{\tabcolsep}{3pt}
\begin{tabular}{l c c l l l c}
\toprule
 &  &  & \multicolumn{2}{c}{real images} & gray canvas & \\
\cmidrule(lr){4-5}\cmidrule(lr){6-6}
run & step & density & level [95\% CI] & $\Delta$ vs untrained [95\% CI] & level [95\% CI] & criteria met \\
\midrule
untrained 7B & 0 & 8-pt & 0.585 [0.500, 0.665] & -- & 0/200 (100 programs) & 4 of 5 \\
untrained 7B & 0 & 12-pt & 0.560 [0.470, 0.645] & -- & 0/200 (100 programs) & 4 of 5 \\
untrained 7B & 0 & 20-pt & 0.425 [0.335, 0.515] & -- & 0/200 (100 programs) & 5 of 5 \\
\cond{Real}, run 1 & 30 & 8-pt & 0.935 [0.895, 0.970] & $+$0.350 [0.275, 0.430] & 0/200 (100 programs) & 4 of 5 \\
\cond{Real}, run 1 & 30 & 12-pt & 0.860 [0.805, 0.910] & $+$0.300 [0.225, 0.380] & 0/200 (100 programs) & 4 of 5 \\
\cond{Real}, run 1 & 30 & 20-pt & 0.740 [0.665, 0.815] & $+$0.315 [0.235, 0.400] & 0/200 (100 programs) & 5 of 5 \\
\cond{Real}, run 1 & 100 & 8-pt & 0.990 [0.975, 1.000] & $+$0.405 [0.325, 0.490] & 0/200 (100 programs) & 4 of 5 \\
\cond{Real}, run 1 & 100 & 12-pt & 0.940 [0.905, 0.970] & $+$0.380 [0.295, 0.470] & 0/200 (100 programs) & 4 of 5 \\
\cond{Real}, run 1 & 100 & 20-pt & 0.875 [0.815, 0.930] & $+$0.450 [0.365, 0.540] & 0/200 (100 programs) & 5 of 5 \\
\cond{Real}, run 2 & 30 & 8-pt & 0.915 [0.865, 0.960] & $+$0.330 [0.250, 0.415] & 0/200 (100 programs) & 4 of 5 \\
\cond{Real}, run 2 & 30 & 12-pt & 0.850 [0.795, 0.905] & $+$0.290 [0.215, 0.370] & 0/200 (100 programs) & 4 of 5 \\
\cond{Real}, run 2 & 30 & 20-pt & 0.720 [0.645, 0.795] & $+$0.295 [0.215, 0.375] & 0/200 (100 programs) & 5 of 5 \\
\cond{Gray}-trained control & 30 & 8-pt & 0.590 [0.510, 0.670] & $+$0.005 [$-$0.035, 0.050] & 0.050 [0.020, 0.090] & 4 of 5 \\
\cond{Gray}-trained control & 30 & 12-pt & 0.580 [0.490, 0.665] & $+$0.020 [$-$0.025, 0.070] & 0.060 [0.025, 0.100] & 4 of 5 \\
\cond{Gray}-trained control & 30 & 20-pt & 0.385 [0.300, 0.475] & $-$0.040 [$-$0.085, 0.000] & 0/200 (100 programs) & 5 of 5 \\
\bottomrule
\end{tabular}
\par\vspace{2pt}
{\scriptsize\raggedright With a gray canvas the \cond{Gray}-trained control changes by $+$0.050 [0.020, 0.090] and $+$0.060 [0.025, 0.100] from the untrained model at 8 and 12 points; every other gray-canvas cell, its 20-point cell included, is 0 of 200: each response is a fixed default outside the answer set.\par}
\end{table}

\begin{table}[h]
\caption{Attribution at matched step 30: the \cond{Real} gain and the \cond{Gray}-trained control's change (both evaluated with real images) and their difference $\tau_{\mathrm{vis}}$ (Eq.~(\ref{eq:gains})), on the development set (dev) and replicated on the confirmatory set (conf); task: discovery (L3) or the identification probe, with each model's gray-canvas count at the same checkpoint (correct/total). 95\% intervals from resampling scene programs, with the same draws for both models.}
\label{tab:attribution}
\centering
\scriptsize
\setlength{\tabcolsep}{3pt}
\begin{tabular}{l c l l l l c c}
\toprule
 &  &  &  &  &  & \multicolumn{2}{c}{gray-canvas count} \\
\cmidrule(lr){7-8}
set & density & task & \cond{Real} $\Delta$ [95\% CI] & control $\Delta$ [95\% CI] & $\tau_{\mathrm{vis}}$ [95\% CI] & \cond{Real} & control \\
\midrule
dev & 8-pt & L3 & $+$0.305 [0.230, 0.380] & $+$0.030 [$-$0.005, 0.065] & $+$0.275 [0.205, 0.350] & 0/200 & 11/200 \\
dev & 8-pt & probe & $+$0.060 [0.025, 0.105] & $+$0.005 [$-$0.020, 0.035] & $+$0.055 [0.020, 0.095] & 0/200 & 0/200 \\
dev & 12-pt & L3 & $+$0.315 [0.230, 0.400] & $-$0.035 [$-$0.080, 0.005] & $+$0.350 [0.265, 0.435] & 0/200 & 9/200 \\
dev & 12-pt & probe & $+$0.065 [0.025, 0.110] & $-$0.005 [$-$0.035, 0.020] & $+$0.070 [0.030, 0.120] & 0/200 & 0/200 \\
dev & 20-pt & L3 & $+$0.330 [0.255, 0.410] & $-$0.030 [$-$0.070, 0.010] & $+$0.360 [0.275, 0.445] & 0/200 & 8/200 \\
dev & 20-pt & probe & $+$0.135 [0.085, 0.190] & $+$0.020 [0.000, 0.045] & $+$0.115 [0.070, 0.170] & 0/200 & 0/200 \\
conf & 8-pt & L3 & $+$0.350 [0.275, 0.430] & $+$0.005 [$-$0.035, 0.050] & $+$0.345 [0.270, 0.425] & 0/200 & 10/200 \\
conf & 8-pt & probe & $+$0.090 [0.045, 0.145] & $+$0.020 [0.005, 0.040] & $+$0.070 [0.025, 0.120] & -- & -- \\
conf & 12-pt & L3 & $+$0.300 [0.225, 0.380] & $+$0.020 [$-$0.025, 0.070] & $+$0.280 [0.210, 0.355] & 0/200 & 12/200 \\
conf & 12-pt & probe & $+$0.080 [0.035, 0.135] & $-$0.005 [$-$0.035, 0.020] & $+$0.085 [0.035, 0.140] & -- & -- \\
conf & 20-pt & L3 & $+$0.315 [0.235, 0.400] & $-$0.040 [$-$0.085, 0.000] & $+$0.355 [0.270, 0.440] & 0/200 & 0/200 \\
conf & 20-pt & probe & $+$0.150 [0.090, 0.215] & $+$0.025 [0.000, 0.050] & $+$0.125 [0.070, 0.185] & -- & -- \\
\bottomrule
\end{tabular}
\end{table}

\begin{table}[h]
\caption{Transfer to the independently constructed grounding tasks (Figure~\ref{fig:flip}(c)) at steps 0, 30 and 100 for both runs: pair accuracy with 95\% intervals and, beneath, the paired change from the untrained model (5,000-draw item bootstrap). Changes are computed from unrounded accuracies, so they can differ by 0.001 from the difference of the rounded levels (as do the labels of Figure~\ref{fig:flip}(c)). Step-30 twin levels are from the evaluation of Table~\ref{tab:seeds}; their changes are differences of rounded levels and carry no interval.}
\label{tab:transfer}
\centering
\scriptsize
\setlength{\tabcolsep}{3pt}
\begin{tabular}{l c >{\raggedright\arraybackslash}p{2.75cm} >{\raggedright\arraybackslash}p{2.75cm} >{\raggedright\arraybackslash}p{2.75cm} >{\raggedright\arraybackslash}p{2.75cm}}
\toprule
run & step & grounding task (600 pairs) & twin (600) & cued-readout task (300) & header-cued table (300) \\
\midrule
untrained 7B & 0 & 0.768 [0.733, 0.802] & 0.728 [0.692, 0.762] & 0.673 [0.620, 0.727] & 0.993 [0.983, 1.000] \\
run 1 & 30 & 0.930 [0.908, 0.950]\newline $\Delta$ $+$0.162 [0.130, 0.193] & 0.900\newline $\Delta$ $+$0.172 & 0.680 [0.627, 0.733]\newline $\Delta$ $+$0.007 [$-$0.023, 0.037] & 0.997 [0.990, 1.000]\newline $\Delta$ $+$0.003 [0.000, 0.010] \\
 & 100 & 0.948 [0.930, 0.965]\newline $\Delta$ $+$0.180 [0.147, 0.215] & 0.937 [0.917, 0.955]\newline $\Delta$ $+$0.208 [0.175, 0.243] & 0.707 [0.657, 0.757]\newline $\Delta$ $+$0.033 [0.000, 0.070] & 0.997 [0.990, 1.000]\newline $\Delta$ $+$0.003 [0.000, 0.010] \\
run 2 & 30 & 0.935 [0.915, 0.955]\newline $\Delta$ $+$0.167 [0.137, 0.198] & 0.905\newline $\Delta$ $+$0.177 & 0.690 [0.640, 0.743]\newline $\Delta$ $+$0.017 [$-$0.013, 0.047] & 0.997 [0.990, 1.000]\newline $\Delta$ $+$0.003 [0.000, 0.010] \\
 & 100 & 0.962 [0.947, 0.977]\newline $\Delta$ $+$0.193 [0.162, 0.227] & -- & 0.727 [0.677, 0.777]\newline $\Delta$ $+$0.053 [0.020, 0.087] & 0.997 [0.990, 1.000]\newline $\Delta$ $+$0.003 [0.000, 0.010] \\
\bottomrule
\end{tabular}
\end{table}

\begin{table}[h]
\caption{The same-scene hierarchy on the 20-point confirmatory scenes, a density never trained (100 scene programs in the top block): member accuracy with real images with its 95\% interval over scene programs, its change from the untrained model, and, for pairs whose answer changes between the twins, the number of scene programs with both members correct. Top: the four tasks on the target-stable and invariance pairs, the population of the discovery result (200 members). Bottom: pair roles separately (100 members each); the cued levels have no target-switch pairs. L1 and L2 ask the same question, and L1 adds only the arrow; L1, L2 and the 20-point density were never trained, while the probe was.}
\label{tab:hierarchy}
\centering
\scriptsize
\setlength{\tabcolsep}{3pt}
\begin{tabular}{l >{\raggedright\arraybackslash}p{2.12cm} >{\raggedright\arraybackslash}p{2.12cm} >{\raggedright\arraybackslash}p{2.12cm} >{\raggedright\arraybackslash}p{2.12cm} >{\raggedright\arraybackslash}p{2.12cm}}
\toprule
task & untrained & \cond{Real} run 1, 30 & \cond{Real} run 2, 30 & \cond{Gray}-trained control, 30 & \cond{Real} run 1, 100 \\
\midrule
L1 cued readout & 0.610 [0.525, 0.695] & 0.820 [0.755, 0.880]\newline $\Delta$ $+$0.210 [0.140, 0.285] & 0.810 [0.745, 0.870]\newline $\Delta$ $+$0.200 [0.130, 0.270] & 0.640 [0.555, 0.720]\newline $\Delta$ $+$0.030 [$-$0.010, 0.075] & 0.890 [0.835, 0.935]\newline $\Delta$ $+$0.280 [0.200, 0.360] \\
L2 find-and-bind & 0.620 [0.530, 0.705] & 0.900 [0.850, 0.945]\newline $\Delta$ $+$0.280 [0.200, 0.360] & 0.875 [0.815, 0.930]\newline $\Delta$ $+$0.255 [0.180, 0.330] & 0.680 [0.600, 0.760]\newline $\Delta$ $+$0.060 [0.020, 0.105] & 0.950 [0.915, 0.980]\newline $\Delta$ $+$0.330 [0.250, 0.415] \\
L3 discovery & 0.425 [0.335, 0.515] & 0.740 [0.665, 0.815]\newline $\Delta$ $+$0.315 [0.235, 0.400] & 0.720 [0.645, 0.795]\newline $\Delta$ $+$0.295 [0.215, 0.375] & 0.385 [0.300, 0.475]\newline $\Delta$ $-$0.040 [$-$0.085, 0.000] & 0.875 [0.815, 0.930]\newline $\Delta$ $+$0.450 [0.365, 0.540] \\
identification probe & 0.825 [0.755, 0.890] & 0.975 [0.940, 1.000]\newline $\Delta$ $+$0.150 [0.090, 0.215] & 0.975 [0.940, 1.000]\newline $\Delta$ $+$0.150 [0.090, 0.215] & 0.850 [0.785, 0.905]\newline $\Delta$ $+$0.025 [0.000, 0.050] & 0.985 [0.960, 1.000]\newline $\Delta$ $+$0.160 [0.100, 0.225] \\
\bottomrule
\end{tabular}
\par\medskip
\begin{tabular}{l >{\raggedright\arraybackslash}p{2.12cm} >{\raggedright\arraybackslash}p{2.12cm} >{\raggedright\arraybackslash}p{2.12cm} >{\raggedright\arraybackslash}p{2.12cm} >{\raggedright\arraybackslash}p{2.12cm}}
\toprule
task & untrained & \cond{Real} run 1, 30 & \cond{Real} run 2, 30 & \cond{Gray}-trained control, 30 & \cond{Real} run 1, 100 \\
\midrule
L3, target-switch & 0.430 [0.320, 0.540]\newline both: 13/50 & 0.780 [0.700, 0.860]\newline $\Delta$ $+$0.350 [0.250, 0.450]\newline both: 30/50 & 0.750 [0.670, 0.830]\newline $\Delta$ $+$0.320 [0.230, 0.420]\newline both: 27/50 & 0.440 [0.340, 0.550]\newline $\Delta$ $+$0.010 [$-$0.060, 0.080]\newline both: 12/50 & 0.880 [0.810, 0.940]\newline $\Delta$ $+$0.450 [0.340, 0.560]\newline both: 39/50 \\
probe, target-switch & 0.890 [0.830, 0.940]\newline both: 39/50 & 0.990 [0.970, 1.000]\newline $\Delta$ $+$0.100 [0.050, 0.160]\newline both: 49/50 & 0.990 [0.970, 1.000]\newline $\Delta$ $+$0.100 [0.050, 0.160]\newline both: 49/50 & 0.910 [0.860, 0.960]\newline $\Delta$ $+$0.020 [$-$0.020, 0.060]\newline both: 41/50 & 100/100 (50 programs)\newline $\Delta$ $+$0.110 [0.060, 0.170]\newline both: 50/50 \\
L3, target-stable & 0.410 [0.290, 0.530]\newline both: 14/50 & 0.730 [0.630, 0.820]\newline $\Delta$ $+$0.320 [0.230, 0.420]\newline both: 29/50 & 0.720 [0.620, 0.820]\newline $\Delta$ $+$0.310 [0.200, 0.420]\newline both: 29/50 & 0.380 [0.270, 0.490]\newline $\Delta$ $-$0.030 [$-$0.090, 0.030]\newline both: 12/50 & 0.890 [0.820, 0.950]\newline $\Delta$ $+$0.480 [0.370, 0.590]\newline both: 40/50 \\
L3, invariance & 0.440 [0.300, 0.580] & 0.750 [0.630, 0.860]\newline $\Delta$ $+$0.310 [0.180, 0.450] & 0.720 [0.600, 0.830]\newline $\Delta$ $+$0.280 [0.160, 0.400] & 0.390 [0.260, 0.520]\newline $\Delta$ $-$0.050 [$-$0.120, 0.010] & 0.860 [0.760, 0.940]\newline $\Delta$ $+$0.420 [0.290, 0.550] \\
\bottomrule
\end{tabular}
\par\vspace{2pt}
{\scriptsize\raggedright On target-switch discovery, \cond{Real} minus the \cond{Gray}-trained control at step 30 is $+$0.340 [0.250, 0.430]. A cell with every member correct is given as its count.\par}
\end{table}

\textbf{Seed replication and the named-target ablation.} Three further independent seeds (1--3) repeat the matched step-30 comparison of \S\ref{sec:drives} with fresh data order and rollout sampling; seed 0 is the original pair (\cond{Real} runs 1 and 2, which share a seed, and the \cond{Gray}-trained control). Each new seed also trains a third condition, the \textbf{named-target corpus}: the same scene programs with the find-and-bind (L2) question in place of the relation-only (L3) question and the identification probe unchanged, so that discovery itself is never trained. Table~\ref{tab:seeds} gives every run at step 30 on the 20-point confirmatory set and on an additional untouched set (333 scene programs per pair role; 1,332 discovery members and 333 target-switch pairs) that had not been used in any development or analysis before this evaluation; Table~\ref{tab:seedcontrasts} summarizes the contrasts over seeds.

The attribution replicates seed by seed. \cond{Real} exceeds \cond{Gray} on confirmatory discovery by +0.354 (seed SD 0.019; every seed at least +0.335) and on the untouched set by +0.339 (SD 0.021); \cond{Gray} changes from the untrained model by $-$0.033 (SD 0.032) and never by more than $-$0.075. Transfer to the grounding task replicates as well: \cond{Real} minus untrained +0.145 (SD 0.016), \cond{Real} minus \cond{Gray} +0.141 (SD 0.029). The step-30 \cond{Real} levels themselves are 0.730--0.760 (mean 0.746, SD 0.014). Seed 1 then continues to step 100 on the seed-0 schedule, with restarts at steps 30, 50 and 75, and reaches 0.895 on confirmatory discovery (seed 0: 0.875), 0.900 on the untouched set, 0.942 and 0.925 on the grounding task and its twin, and 0.905 and 0.960 on cued readout and find-and-bind (identification probe: 297 of 300). The two step-100 endpoints are reported directly, without a two-seed interval. The named-target corpus reaches 0.695--0.720 on confirmatory discovery without ever training it, +0.042 below the relation-only corpus (seed-level 95\% interval [$-$0.027, +0.110]; +0.045 [$-$0.005, +0.095] on the untouched set), and transfers to the grounding task about as strongly (seed means 0.928 against 0.907). A model taught to find a named point and to name the point a relation picks out can, without further instruction, find the point a relation picks out.

Confirmatory discovery on gray test images is 0 of 200 for every run. Without the image, per-twin accuracy on the untouched set is 0.069--0.078 for every model, the untrained one included; the contrast between \cond{Real} and \cond{Gray} is unaffected, since both conditions share that floor.

\begin{table}[h]
\caption{Seed replication (\S\S\ref{sec:acquired}--\ref{sec:transfer}): seeds 1--3 of the \cond{Real}, \cond{Gray}-trained and named-target conditions at step 30, and the \cond{Real} run of seed 1 continued to step 100, beside seed 0 (Tables~\ref{tab:confirmatory}--\ref{tab:hierarchy}). Discovery member accuracy on the 20-point confirmatory set (200 members) and on the untouched set (1,332 members); pair accuracy on the grounding task and its twin; target-switch discovery with both members correct on the confirmatory (of 50 pairs) and untouched (of 333) sets; find-and-bind (200 members) on the confirmatory scenes. Counts are given where recorded.}
\label{tab:seeds}
\centering
\scriptsize
\setlength{\tabcolsep}{3pt}
\begin{tabular}{l >{\raggedright\arraybackslash}p{40pt} >{\raggedright\arraybackslash}p{46pt} >{\raggedright\arraybackslash}p{56pt} >{\raggedleft\arraybackslash}p{30pt} r >{\raggedleft\arraybackslash}p{26pt} >{\raggedright\arraybackslash}p{46pt} >{\raggedleft\arraybackslash}p{26pt}}
\toprule
seed & condition & discovery, confirmatory & discovery, untouched & grounding task & twin & target-switch, conf. & target-switch, untouched & find-and-bind \\
\midrule
-- & untrained 7B & 0.425 (85/200) & 0.471 (628/1,332) & 0.768 & 0.728 & 13/50 & 0.318 & 0.620 \\
0 & \cond{Real}, run 1 & 0.740 & 0.791 & 0.930 & 0.900 & 30/50 & 0.586 & 0.900 \\
0 & \cond{Real}, run 2 (same seed) & 0.720 & 0.770 & 0.935 & 0.905 & 27/50 & 0.562 & 0.875 \\
0 & \cond{Gray}-trained & 0.385 & 0.440 & 0.770 & 0.730 & 12/50 & 0.252 & 0.680 \\
1 & \cond{Real} & 0.760 (152/200) & 0.792 (1,055/1,332) & 0.907 & 0.885 & 30/50 & 0.583 & 0.915 \\
1 & \cond{Real}, step~100 & 0.895 (179/200) & 0.900 (1,199/1,332) & 0.942 & 0.925 & -- & -- & 0.960 \\
1 & \cond{Gray}-trained & 0.415 (83/200) & 0.451 (601/1,332) & 0.778 & 0.742 & 13/50 & 0.267 (89/333) & 0.685 \\
1 & named-target & 0.705 (141/200) & 0.725 (966/1,332) & 0.943 & 0.912 & 24/50 & 0.502 (167/333) & 0.960 \\
2 & \cond{Real} & 0.730 (146/200) & 0.769 (1,024/1,332) & 0.922 & 0.898 & 26/50 & 0.544 (181/333) & 0.925 \\
2 & \cond{Gray}-trained & 0.350 (70/200) & 0.414 (551/1,332) & 0.752 & 0.715 & 10/50 & 0.213 (71/333) & 0.645 \\
2 & named-target & 0.720 (144/200) & 0.742 (988/1,332) & 0.903 & 0.873 & 29/50 & 0.529 (176/333) & 0.980 \\
3 & \cond{Real} & 0.755 (151/200) & 0.762 (1,015/1,332) & 0.893 & 0.882 & 33/50 & 0.595 (198/333) & 0.890 \\
3 & \cond{Gray}-trained & 0.420 (84/200) & 0.453 (603/1,332) & 0.787 & 0.737 & 14/50 & 0.282 (94/333) & 0.705 \\
3 & named-target & 0.695 (139/200) & 0.720 (959/1,332) & 0.938 & 0.923 & 25/50 & 0.508 (169/333) & 0.955 \\
\bottomrule
\end{tabular}
\end{table}

\begin{table}[h]
\caption{Levels and contrasts over seeds at step 30 (seed 0 uses \cond{Real} run 1): per-seed values, their mean and sample standard deviation, and, for the contrasts used as seed-level claims in \S\S\ref{sec:acquired}--\ref{sec:transfer}, the range and the $t$-based 95\% interval over seeds (Appendix~\ref{app:stats}).}
\label{tab:seedcontrasts}
\centering
\scriptsize
\setlength{\tabcolsep}{3pt}
\begin{tabular}{>{\raggedright\arraybackslash}p{54pt} >{\raggedright\arraybackslash}p{44pt} c l r r r >{\raggedright\arraybackslash}p{53pt}}
\toprule
quantity & endpoint & seeds & per-seed values & mean & seed SD & range & seed-level 95\% interval \\
\midrule
\cond{Real} level & discovery, confirmatory & 0--3 & 0.740, 0.760, 0.730, 0.755 & 0.746 & 0.014 & 0.030 & -- \\
\cond{Real} $-$ \cond{Gray} & discovery, confirmatory & 0--3 & $+$0.355, $+$0.345, $+$0.380, $+$0.335 & $+$0.354 & 0.019 & 0.045 & [$+$0.323, $+$0.384] \\
\cond{Real} $-$ \cond{Gray} & discovery, untouched & 0--3 & $+$0.351, $+$0.341, $+$0.355, $+$0.309 & $+$0.339 & 0.021 & 0.046 & [$+$0.306, $+$0.372] \\
\cond{Real} $-$ untrained & discovery, confirmatory & 0--3 & $+$0.315, $+$0.335, $+$0.305, $+$0.330 & $+$0.321 & 0.014 & -- & -- \\
\cond{Real} $-$ untrained & discovery, untouched & 0--3 & $+$0.320, $+$0.321, $+$0.297, $+$0.291 & $+$0.307 & 0.015 & -- & -- \\
\cond{Gray} $-$ untrained & discovery, confirmatory & 0--3 & $-$0.040, $-$0.010, $-$0.075, $-$0.005 & $-$0.033 & 0.032 & -- & -- \\
\cond{Real} $-$ untrained & grounding task & 0--3 & $+$0.162, $+$0.139, $+$0.154, $+$0.125 & $+$0.145 & 0.016 & -- & -- \\
\cond{Real} $-$ \cond{Gray} & grounding task & 0--3 & $+$0.160, $+$0.129, $+$0.170, $+$0.106 & $+$0.141 & 0.029 & -- & -- \\
relation-only $-$ named-target & discovery, untouched & 1--3 & $+$0.067, $+$0.027, $+$0.042 & $+$0.045 & 0.020 & 0.040 & [$-$0.005, $+$0.095] \\
relation-only $-$ named-target & discovery, confirmatory & 1--3 & $+$0.055, $+$0.010, $+$0.060 & $+$0.042 & 0.028 & 0.050 & [$-$0.027, $+$0.110] \\
\bottomrule
\end{tabular}
\end{table}

\textbf{Composite operations.} Discovery requires identifying the target and then locating and reading it. Table~\ref{tab:composite} splits the 200 confirmatory members by whether the untrained model answers the identification probe for that member correctly (165 members) or not (35) and reports step-30 discovery within each subset for seed 0. \cond{Real} raises discovery in both: from 0.485 to 0.764 where identification was already available, and from 0.143 to 0.629 where it was not. \cond{Gray} leaves the second subset unchanged and the first slightly lower. The gain is not confined to targets the base model could already identify; identification is itself acquired, which is what the compositional reading of the named-target ablation requires. The split conditions on a base-model outcome and is reported descriptively, without intervals.

\begin{table}[h]
\caption{Step-30 discovery on the 20-point confirmatory set (200 members; seed 0, \cond{Real} run 1 and the \cond{Gray}-trained control), by whether the untrained model's identification probe for the same member is correct.}
\label{tab:composite}
\centering
\scriptsize
\setlength{\tabcolsep}{3pt}
\begin{tabular}{l r r r r}
\toprule
subset & members & untrained & \cond{Real} & \cond{Gray}-trained \\
\midrule
probe correct & 165 & 0.485 (80/165) & 0.764 (126/165) & 0.436 (72/165) \\
probe wrong & 35 & 0.143 (5/35) & 0.629 (22/35) & 0.143 (5/35) \\
all & 200 & 0.425 (85/200) & 0.740 (148/200) & 0.385 (77/200) \\
\bottomrule
\end{tabular}
\end{table}

\textbf{Answers available without looking.} The \cond{Gray}-trained control removes the image and with it almost all correctness reward (Table~\ref{tab:seedrewards}). A third condition keeps the image and the reward and adds a text shortcut: the question-blind Qwen2.5-VL-72B caption of each training image (the captions of criterion 4, Appendix~\ref{app:ladder}) is placed in the prompt after the image and before the question, with the 2,880 prompts, reward, budget, starting checkpoint and data-order and rollout seeds of \cond{Real} seeds 1--3 otherwise unchanged. At the training densities the caption alone answers 0.580 (8 points) and 0.445 (12 points) of discovery questions, so the answer is often available without looking; at 20 points it is not (Table~\ref{tab:criteria}). Table~\ref{tab:shortcut} gives the three seeds at step 30, tested with the image alone and with the image and caption; Table~\ref{tab:shortcutcontrasts} gives the contrasts over seeds.

Reward is earned as under \cond{Real}: the accuracy component starts at 0.52--0.57 (\cond{Real}: 0.29--0.34) and reaches 0.97--0.99 by step 30 (Table~\ref{tab:seedrewards}). The skill is not. With the image alone, confirmatory discovery is 0.550, 0.515 and 0.570, a gain of +0.120 over the untrained model against +0.323 for \cond{Real}; every shortcut seed is below every \cond{Real} seed, with a seed-level gap of +0.203 [+0.163, +0.243]. The untouched set shows the same gap (+0.195 [+0.132, +0.258]), and grounding transfer is +0.043 against +0.125 to +0.154 for \cond{Real}. With its caption at test the arm reaches 0.610--0.655, still below every \cond{Real} seed's image-only level, so the gap is not a change of input format; what the arm learned was to read the caption. Gray test images give 0 of 200 in every seed.

\begin{table}[h]
\caption{The shortcut-accessible twin at step 30, seeds 1--3, tested with the image alone (the paper's evaluation) and with the image and its caption (the training input): discovery member accuracy on the confirmatory (200) and untouched (1,332) sets, pair accuracy on the grounding task (600) and discovery with gray test images. \cond{Real} and \cond{Gray}-trained seeds 1--3 and the untrained model are in Table~\ref{tab:seeds}.}
\label{tab:shortcut}
\centering
\scriptsize
\setlength{\tabcolsep}{3pt}
\begin{tabular}{c l l l l l}
\toprule
seed & test input & discovery, confirmatory & discovery, untouched & grounding task & gray test images \\
\midrule
1 & image only & 0.550 (110/200) & 0.589 (784/1,332) & 0.820 (492/600) & 0/200 \\
2 & image only & 0.515 (103/200) & 0.553 (737/1,332) & 0.802 (481/600) & 0/200 \\
3 & image only & 0.570 (114/200) & 0.595 (793/1,332) & 0.812 (487/600) & 0/200 \\
1 & image and caption & 0.645 (129/200) & 0.669 (891/1,332) & -- & -- \\
2 & image and caption & 0.610 (122/200) & 0.646 (861/1,332) & -- & -- \\
3 & image and caption & 0.655 (131/200) & 0.682 (908/1,332) & -- & -- \\
\bottomrule
\end{tabular}
\end{table}

\begin{table}[h]
\caption{Contrasts over seeds for the shortcut-accessible twin (image only), in the format of Table~\ref{tab:seedcontrasts}; \cond{Real} values from Table~\ref{tab:seedcontrasts} for comparison.}
\label{tab:shortcutcontrasts}
\centering
\scriptsize
\setlength{\tabcolsep}{3pt}
\begin{tabular}{>{\raggedright\arraybackslash}p{54pt} >{\raggedright\arraybackslash}p{44pt} l r r >{\raggedright\arraybackslash}p{53pt} >{\raggedright\arraybackslash}p{76pt}}
\toprule
quantity & endpoint & per-seed values (1, 2, 3) & mean & seed SD & seed-level 95\% interval & \cond{Real} reference \\
\midrule
shortcut level & discovery, confirmatory & 0.550, 0.515, 0.570 & 0.545 & 0.028 & [0.476, 0.614] & 0.760, 0.730, 0.755 \\
shortcut $-$ untrained & discovery, confirmatory & $+$0.125, $+$0.090, $+$0.145 & $+$0.120 & 0.028 & [$+$0.051, $+$0.189] & $+$0.335, $+$0.305, $+$0.330 \\
\cond{Real} $-$ shortcut & discovery, confirmatory & $+$0.210, $+$0.215, $+$0.185 & $+$0.203 & 0.016 & [$+$0.163, $+$0.243] & \cond{Real} $-$ \cond{Gray}: $+$0.345, $+$0.380, $+$0.335 \\
shortcut level & discovery, untouched & 0.589, 0.553, 0.595 & 0.579 & 0.023 & [0.523, 0.635] & 0.792, 0.769, 0.762 \\
\cond{Real} $-$ shortcut & discovery, untouched & $+$0.203, $+$0.215, $+$0.167 & $+$0.195 & 0.025 & [$+$0.132, $+$0.258] & \cond{Real} $-$ \cond{Gray}: $+$0.341, $+$0.355, $+$0.309 \\
shortcut $-$ untrained & grounding task & $+$0.052, $+$0.033, $+$0.043 & $+$0.043 & 0.009 & [$+$0.020, $+$0.066] & $+$0.139, $+$0.154, $+$0.125 \\
\bottomrule
\end{tabular}
\end{table}

\AppendixBarrier
\section{Dose, Audit Standardization, and Capacity}\label{app:dose}

\textbf{Dose response.} Table~\ref{tab:dose} gives all four mixture conditions of \S\ref{sec:dose}. Each grid point is a single run from the common start checkpoint with the same reward, budget (30 steps, 240 prompts per step, five responses per prompt) and stream definition; the constructed prompts per step are 0, 80, 160 and 240. On the never-trained density the two intermediate mixtures are tied (+0.275 and +0.270), and the fully constructed condition gives the largest gain on every density. On this fixed-budget grid the audited mass $m(f)$ is nearly linear in $f$ (correlation 0.997; largest residual from a straight line 0.006) and serves as the audit's descriptor of each mixture.

\textbf{Audit association and equal exposure.} Table~\ref{tab:slopes} reports the association between step-30 accuracy and the audited mass, raw and standardized to a common initial-success profile (Appendix~\ref{app:standardization}); the raw slope depends strongly on the condition without constructed data, and the standardized mass is itself nonlinear in $f$. It also gives both equal-exposure comparisons: at equal constructed-prompt exposure the one-third mixture is ahead of the fully constructed run on all three densities, and the two-thirds mixture is within 0.035 of it (the 12- and 20-point intervals include zero). Its last block compares the mixtures directly at step 30.

\textbf{Ordinary-task accuracy.} On the decontaminated held-out ViRL39K split (4,239 items, none shared with any training stream), greedy step-30 accuracy with real images is 0.548, 0.519, 0.522 and 0.419 for $f=0,\tfrac13,\tfrac23,1$, each at least 0.20 above the same model's image-removed accuracy. The fully constructed run's deficit lies on the 1,215 multiple-choice items (0.325, against 0.649--0.660 for the other three conditions); on numeric and free-text items it is within 0.05 of the other conditions. The untrained model scores 0.509 with images (2,157 of 4,239; 0.250 without) and 0.640 on the multiple-choice items, so the two mixtures match it and ordinary-only training exceeds it. Table~\ref{tab:mixbench} extends the comparison to two audited benchmarks: with images, the one-third mixture matches the untrained model on MathVista testmini (0.696 against 0.682) and BLINK (0.547 against 0.536), with blind retention unchanged; the fully constructed run falls on both.

\textbf{Capacity.} Table~\ref{tab:capacity} gives the capacity comparison of \S\ref{sec:capacity} per checkpoint.

\begin{table}[h]
\caption{Dose response at step 30 by constructed fraction $f$ (same 7B starting model, reward, recipe and 7,200 prompt presentations in all four conditions): audited mass $m(f)$ (Eq.~(\ref{eq:mass})), standardized mass $\widetilde m(f)$ (Appendix~\ref{app:standardization}) and the share of audit-joined training rows with $0<b_i<1$ (in-band); per density, discovery accuracy with real images and its gain over the untrained model with a 95\% interval from resampling scene programs; and grounding-task pair accuracy. With a gray canvas, discovery accuracy is 0 of 200 in every cell of every mixture.}
\label{tab:dose}
\centering
\scriptsize
\setlength{\tabcolsep}{3pt}
\begin{tabular}{c r r r >{\raggedright\arraybackslash}p{2.4cm} >{\raggedright\arraybackslash}p{2.4cm} >{\raggedright\arraybackslash}p{2.4cm} l}
\toprule
$f$ & $m(f)$ & $\widetilde m(f)$ & in-band & 8-pt: level; gain [95\% CI] & 12-pt & 20-pt (never trained) & grounding [95\% CI] \\
\midrule
0 & 0.126 & 0.094 & 0.752 & 0.740\newline $+$0.080 [0.030, 0.135] & 0.570\newline $-$0.005 [$-$0.035, 0.030] & 0.460\newline $-$0.010 [$-$0.040, 0.025] & 0.797 [0.763, 0.828] \\
$\tfrac13$ & 0.165 & 0.229 & 0.900 & 0.935\newline $+$0.275 [0.200, 0.355] & 0.805\newline $+$0.230 [0.155, 0.305] & 0.745\newline $+$0.275 [0.205, 0.350] & 0.908 [0.885, 0.930] \\
$\tfrac23$ & 0.199 & 0.240 & 0.926 & 0.950\newline $+$0.290 [0.215, 0.370] & 0.850\newline $+$0.275 [0.195, 0.360] & 0.740\newline $+$0.270 [0.200, 0.345] & 0.927 [0.905, 0.947] \\
1 & 0.249 & 0.251 & -- & 0.965\newline $+$0.305 [0.230, 0.380] & 0.890\newline $+$0.315 [0.230, 0.400] & 0.800\newline $+$0.330 [0.255, 0.410] & 0.930 [0.908, 0.950] \\
\bottomrule
\end{tabular}
\par\vspace{2pt}
{\scriptsize\raggedright The in-band share was not recorded for $f=1$: the stream profiles cover only the three conditions that contain ordinary data.\par}
\end{table}

\begin{table}[h]
\caption{Audit association, equal-exposure and direct comparisons for the mixtures. Top: ordinary least-squares slope (with intercept, over the four conditions) of step-30 held-out accuracy on the audited mass $m(f)$ and on the standardized mass $\widetilde m(f)$ (Appendix~\ref{app:standardization}); 95\% intervals resample scene programs (grounding pairs for the grounding task) with the audited masses held fixed (Appendix~\ref{app:stats}). Middle: equal constructed-prompt exposure. Discovery accuracy of a mixture at step 30 and of the fully constructed run at the step where it had seen the same number of constructed prompt presentations (five responses each), with their difference and its 95\% interval. The two runs of each comparison differ in total presentations, so these comparisons match constructed exposure, not total training. Bottom: direct comparisons of the mixtures at step 30.}
\label{tab:slopes}
\centering
\scriptsize
\setlength{\tabcolsep}{3pt}
\begin{tabular}{l l l}
\toprule
endpoint & slope on $m(f)$ [95\% CI] & slope on $\widetilde m(f)$ [95\% CI] \\
\midrule
8-pt discovery & 1.670 [1.187, 2.190] & 1.441 [1.005, 1.907] \\
12-pt discovery & 2.446 [1.813, 3.088] & 1.942 [1.432, 2.471] \\
20-pt discovery (never trained) & 2.482 [1.886, 3.083] & 2.079 [1.565, 2.596] \\
grounding task & 1.009 [0.787, 1.233] & 0.861 [0.669, 1.054] \\
\bottomrule
\end{tabular}
\par\medskip
\begin{tabular}{l r r l r r l}
\toprule
 & \multicolumn{3}{c}{$f=\tfrac13$ at step 30 vs.\ $f=1$ at step 10} & \multicolumn{3}{c}{$f=\tfrac23$ at step 30 vs.\ $f=1$ at step 20} \\
\cmidrule(lr){2-4}\cmidrule(lr){5-7}
constructed prompt presentations & 2,400 & 2,400 &  & 4,800 & 4,800 &  \\
total prompt presentations & 7,200 & 2,400 &  & 7,200 & 4,800 &  \\
\midrule
density & mixture & $f=1$ & difference [95\% CI] & mixture & $f=1$ & difference [95\% CI] \\
\midrule
8-pt & 0.935 & 0.870 & $+$0.065 [0.030, 0.105] & 0.950 & 0.945 & $+$0.005 [0.000, 0.015] \\
12-pt & 0.805 & 0.760 & $+$0.045 [0.015, 0.080] & 0.850 & 0.815 & $+$0.035 [$-$0.010, 0.085] \\
20-pt (never trained) & 0.745 & 0.675 & $+$0.070 [0.030, 0.115] & 0.740 & 0.760 & $-$0.020 [$-$0.060, 0.015] \\
\bottomrule
\end{tabular}
\par\medskip
\begin{tabular}{l l l l}
\toprule
density & $f=\tfrac23$ minus $f=\tfrac13$ [95\% CI] & $f=1$ minus $f=\tfrac13$ [95\% CI] & $f=1$ minus $f=\tfrac23$ [95\% CI] \\
\midrule
8-pt & $+$0.015 [0.000, 0.035] & $+$0.030 [0.010, 0.055] & $+$0.015 [0.000, 0.035] \\
12-pt & $+$0.045 [0.000, 0.095] & $+$0.085 [0.045, 0.130] & $+$0.040 [0.000, 0.080] \\
20-pt (never trained) & $-$0.005 [$-$0.050, 0.045] & $+$0.055 [0.025, 0.085] & $+$0.060 [0.020, 0.100] \\
\bottomrule
\end{tabular}
\end{table}

\begin{table}[h]
\caption{Capacity (Figure~\ref{fig:dial}(b)): the 3B and 7B models trained on the same constructed corpus, per checkpoint and density (development set): discovery with real images and its change from the same model untrained, with 95\% intervals from resampling scene programs; the identification probe with real images and with a gray canvas; discovery with a gray canvas; and grounding-task pair accuracy where evaluated. With a gray canvas the probe is exactly zero for both models at every checkpoint, while the 3B model's discovery accuracy is small and nonzero at every checkpoint, including before training: it answers with a fixed coordinate that is sometimes correct.}
\label{tab:capacity}
\centering
\scriptsize
\setlength{\tabcolsep}{3pt}
\begin{tabular}{l c c l l r r r l}
\toprule
 &  &  & \multicolumn{2}{c}{discovery, real images} & \multicolumn{2}{c}{identification probe} & discovery & \\
\cmidrule(lr){4-5}\cmidrule(lr){6-7}
model & step & density & level [95\% CI] & $\Delta$ vs untrained [95\% CI] & real & gray canvas & gray canvas & grounding task [95\% CI] \\
\midrule
3B & 0 & 8-pt & 0.330 [0.255, 0.405] & -- & 0.705 & 0.000 & 0.080 & 0.455 [0.415, 0.495] \\
 &  & 12-pt & 0.260 [0.190, 0.340] & -- & 0.680 & 0.000 & 0.025 &  \\
 &  & 20-pt & 0.245 [0.175, 0.320] & -- & 0.630 & 0.000 & 0.055 &  \\
 & 10 & 8-pt & 0.475 [0.395, 0.555] & $+$0.145 [0.065, 0.225] & 0.875 & 0.000 & 0.080 & -- \\
 &  & 12-pt & 0.425 [0.340, 0.510] & $+$0.165 [0.085, 0.245] & 0.755 & 0.000 & 0.025 &  \\
 &  & 20-pt & 0.320 [0.240, 0.400] & $+$0.075 [$-$0.010, 0.160] & 0.740 & 0.000 & 0.055 &  \\
 & 20 & 8-pt & 0.750 [0.675, 0.820] & $+$0.420 [0.325, 0.510] & 0.960 & 0.000 & 0.105 & -- \\
 &  & 12-pt & 0.635 [0.555, 0.715] & $+$0.375 [0.285, 0.465] & 0.870 & 0.000 & 0.030 &  \\
 &  & 20-pt & 0.465 [0.375, 0.555] & $+$0.220 [0.120, 0.315] & 0.880 & 0.000 & 0.050 &  \\
 & 30 & 8-pt & 0.865 [0.810, 0.915] & $+$0.535 [0.450, 0.620] & 0.985 & 0.000 & 0.060 & 0.585 [0.547, 0.625] \\
 &  & 12-pt & 0.730 [0.655, 0.800] & $+$0.470 [0.375, 0.560] & 0.945 & 0.000 & 0.040 &  \\
 &  & 20-pt & 0.545 [0.460, 0.630] & $+$0.300 [0.200, 0.395] & 0.960 & 0.000 & 0.070 &  \\
7B & 0 & 8-pt & 0.660 [0.580, 0.740] & -- & 0.940 & 0.000 & 0.000 & 0.768 [0.733, 0.802] \\
 &  & 12-pt & 0.575 [0.485, 0.660] & -- & 0.910 & 0.000 & 0.000 &  \\
 &  & 20-pt & 0.470 [0.385, 0.555] & -- & 0.840 & 0.000 & 0.000 &  \\
 & 10 & 8-pt & 0.870 [0.815, 0.920] & $+$0.210 [0.145, 0.275] & 0.985 & 0.000 & 0.000 & -- \\
 &  & 12-pt & 0.760 [0.685, 0.830] & $+$0.185 [0.115, 0.260] & 0.950 & 0.000 & 0.000 &  \\
 &  & 20-pt & 0.675 [0.595, 0.750] & $+$0.205 [0.135, 0.280] & 0.940 & 0.000 & 0.000 &  \\
 & 20 & 8-pt & 0.945 [0.905, 0.980] & $+$0.285 [0.215, 0.360] & 0.995 & 0.000 & 0.000 & -- \\
 &  & 12-pt & 0.815 [0.745, 0.875] & $+$0.240 [0.165, 0.320] & 0.960 & 0.000 & 0.000 &  \\
 &  & 20-pt & 0.760 [0.690, 0.825] & $+$0.290 [0.215, 0.370] & 0.975 & 0.000 & 0.000 &  \\
 & 30 & 8-pt & 0.965 [0.935, 0.990] & $+$0.305 [0.230, 0.380] & 1.000 & 0.000 & 0.000 & 0.930 [0.908, 0.950] \\
 &  & 12-pt & 0.890 [0.830, 0.945] & $+$0.315 [0.230, 0.400] & 0.975 & 0.000 & 0.000 &  \\
 &  & 20-pt & 0.800 [0.730, 0.865] & $+$0.330 [0.255, 0.410] & 0.975 & 0.000 & 0.000 &  \\
\bottomrule
\end{tabular}
\end{table}

\begin{table}[h]
\caption{The untrained 7B model and the step-30 $f=1$ and $f=\tfrac13$ checkpoints on MathVista testmini (1,000 items, 539 of them multiple-choice) and BLINK (1,901 multiple-choice items), with and without the image, with the chance reference $c$ and the chance-corrected blind retention $\rho_{\mathrm{blind}}$ of \S\ref{sec:necessity}. Retention is reported unclipped; without the image the $f=1$ model answers BLINK below chance, so its retention is negative.}
\label{tab:mixbench}
\centering
\scriptsize
\setlength{\tabcolsep}{3pt}
\begin{tabular}{l l l l r r}
\toprule
benchmark & model & with image & no image & $c$ & $\rho_{\mathrm{blind}}$ \\
\midrule
MathVista testmini & untrained 7B & 0.682 (682/1,000) & 0.366 (366/1,000) & 0.179 & 0.372 \\
MathVista testmini & $f=1$, step 30 & 0.532 (532/1,000) & 0.258 (258/1,000) & 0.179 & 0.224 \\
MathVista testmini & $f=\tfrac13$, step 30 & 0.696 (696/1,000) & 0.367 (367/1,000) & 0.179 & 0.364 \\
BLINK & untrained 7B & 0.536 (1,018/1,901) & 0.416 (791/1,901) & 0.377 & 0.247 \\
BLINK & $f=1$, step 30 & 0.457 (869/1,901) & 0.363 (690/1,901) & 0.377 & $-$0.175 \\
BLINK & $f=\tfrac13$, step 30 & 0.547 (1,039/1,901) & 0.420 (798/1,901) & 0.377 & 0.252 \\
\bottomrule
\end{tabular}
\end{table}

\section{Scoring and Uncertainty}\label{app:stats}
\textbf{Answer matching.} A response is correct when its extracted answer is equivalent to the gold under the answer matcher, and trained and untrained models are always scored with the same matcher. One exception to the current matcher revision affects interpretation: the shared-error sets of Table~\ref{tab:overlap} were defined with the earlier revision, under which the untrained 3B model solves 283 of the 600 grounding pairs rather than 273. The current revision changes member accuracy by at most 0.043 per run relative to it, and in a conservative direction: image-removed accuracies fall, pair accuracies without the image stay at exactly 0.000, and gains over the base model rise rather than fall.

\textbf{Units and aggregation.} The grounding suite is scored by pair accuracy (Eq.~(\ref{eq:pairacc}): both members correct); the constructed discovery tasks and their probe by member accuracy. Results over seeds are means of per-seed values, and recovery shares divide seed-averaged gains (Appendix~\ref{app:measures}).

\textbf{Intervals.} Plotted and tabulated intervals are 95\% intervals. Each resamples the evaluation unit of its task (the scene program, the grounding pair or the benchmark item) with the trained models held fixed, and Table~\ref{tab:runbasis} names that unit and the run basis for each result.

\emph{Constructed coordinate tasks} (discovery, the cued levels and the probe; Figures~\ref{fig:flip}(a, b) and~\ref{fig:dial}; Tables~\ref{tab:devgrid}, \ref{tab:confirmatory}, \ref{tab:attribution}, \ref{tab:dose}, \ref{tab:slopes}, \ref{tab:capacity} and~\ref{tab:hierarchy}): the scene program, which contributes both twins, is the independent unit. Levels, gains and contrasts resample scene programs with replacement within their pair role, apply the same draws to every compared model and keep the point-estimate weighting (10,000 draws). A cell whose draws are all equal, such as 0 of 200 correct, is reported as its count and number of scene programs, without an interval.

\emph{Grounding suite and benchmarks}: each grounding pair is its own scene and each benchmark item its own question, so pairs or items are resampled, paired across the compared models: 5,000 draws for the constructed-corpus transfer and the Geometry3K access results (Tables~\ref{tab:transfer} and~\ref{tab:geo7b}), 2,000 for the 7B access pair on the grounding suite and the long-horizon benchmark accuracy, and 10,000 for the long-horizon grounding columns of Table~\ref{tab:trajectories}, the grounding endpoint of the slopes in Table~\ref{tab:slopes}, the extraction split of Table~\ref{tab:extract} and the audit of Appendix~\ref{app:audit}. Figure~\ref{fig:necessity}(a) plots three-seed mean gains with such intervals for cued readout and grounding, and two-seed point estimates without intervals for discovery; Figure~\ref{fig:flip} plots the two runs separately. In Table~\ref{tab:access} the 7B share carries a bootstrap interval and the three-seed 3B shares are point estimates, with per-seed values in Table~\ref{tab:table1perseed}.

\emph{Other quantities:} the identical-wrong-answer rates of Table~\ref{tab:overlap} carry Wilson intervals, and the overlap of the degraded sets is compared with a 10,000-draw permutation null. The slopes of Table~\ref{tab:slopes} hold the audited masses fixed.

\emph{Seed-level intervals} (Table~\ref{tab:seedcontrasts}): for a contrast measured on $n$ independent seeds, the interval is the mean $\pm\, t_{n-1,0.975}\,\mathrm{SD}/\sqrt{n}$, with SD the sample standard deviation over seeds ($n=4$ for \cond{Real} against \cond{Gray}, $n=3$ for the named-target ablation). Seeds, not evaluation items, are the unit; the item-resampled intervals of seed 0 remain those of Tables~\ref{tab:confirmatory}--\ref{tab:hierarchy}.

\textbf{Equivalence test.} The grounding-task gain of \cond{Real} in \S\ref{sec:ladder} is compared with the $\pm0.05$ band by two one-sided tests~\citep{schuirmann1987comparison}; the band was fixed before pilot training, and the test procedure was chosen afterwards. The test uses the 90\% interval of the seed-averaged paired difference, from a 20,000-draw bootstrap over the 600 pairs: [$-$0.003, +0.040], inside the band. It is distinct from the 95\% interval plotted in Figure~\ref{fig:necessity}(a), [$-$0.007, +0.044].

\AppendixBarrier
\section{Mathematical Foundations and Audit Comparison}\label{app:math}
\subsection{Answer-changing paired evaluation}\label{app:pairs}

For pair $j$, write the inputs as $(I_{j,a},q_j)$ and $(I_{j,b},q_j)$, and let $\mathcal A_{j,a}$ and $\mathcal A_{j,b}$ be the sets of extracted answers accepted for its two members. For an answer-changing pair these sets are disjoint: numerical or symbolic equivalence is allowed within each set, but no response matches both gold answers.

\begin{proposition}[image-independent greedy predictors have zero pair accuracy]\label{prop:pairzero}
Suppose every evaluated pair has the same nonvisual input on its two members, disjoint accepted-answer sets, and a fixed deterministic decoding procedure. If the predictor's extracted answer is independent of the image, its pair accuracy in Eq.~(\ref{eq:pairacc}) is zero.
\end{proposition}

\begin{proof}
On each pair, such a predictor returns one answer $u_j$ on both members. Hence $C_{j,a}C_{j,b}=\mathbf 1\{u_j\in\mathcal A_{j,a}\cap\mathcal A_{j,b}\}=0$. Averaging these products gives the claim.
\end{proof}

The proposition concerns answer-changing pair accuracy. It does not describe the member accuracy of the constructed discovery tasks or the target-stable identification probe, whose accuracy without the image is measured directly (\S\ref{sec:drives}).

\subsection{Audit and correctness feedback}\label{app:feedback}

For item $i$, let $C^{\mathrm{image}}_{ik}$ and $C^{\mathrm{none}}_{ik}$ be the binary outcomes of the training reward's accuracy check on audit response $k$, with $K=16$ responses in each condition:
\begin{equation}
b_i=\frac1K\sum_{k=1}^K C^{\mathrm{image}}_{ik},
\qquad
q_{\mathrm{blind},i}=\frac1K\sum_{k=1}^K C^{\mathrm{none}}_{ik},
\qquad
\Delta q_i=b_i-q_{\mathrm{blind},i}.
\label{eq:auditdef}
\end{equation}
These are empirical rates, and the signed difference $\Delta q_i$ is kept without clipping. Let $p$ denote the underlying probability that a sampled response is correct at a fixed prompt and checkpoint; $b_i$ estimates it for the base model. A group of $G$ independent responses is unmixed only when every response is correct or every response is wrong, so it contains both outcomes with probability
\begin{equation}
P_{\mathrm{mix}}(p;G)=1-p^G-(1-p)^G,
\label{eq:pmix}
\end{equation}
which is positive exactly when $0<p<1$ and largest at $p=1/2$, where it equals $1-2^{1-G}$ ($15/16$ for the training groups, $G=5$). With $K=16$ the same expression is the probability that the audit observes both outcomes, $0<b_i<1$; that is a sample event, not exact knowledge of $p$.

Write the reward of response $j$ as $r_j=\tfrac12C_j+\tfrac12F_j$, with $F_j$ the format indicator. With bars denoting means within a prompt's group,
\begin{equation}
r_j-\bar r=\tfrac12(C_j-\bar C)+\tfrac12(F_j-\bar F).
\label{eq:rewarddecomp}
\end{equation}
When format is constant within a group, correctness variation supplies the reward contrast; format variation can also contribute. $P_{\mathrm{mix}}$ therefore measures an opportunity for accuracy-based feedback, not a guarantee of a nonzero GRPO update.

Table~\ref{tab:learnsignal} records this at the start of training for the seeds of Table~\ref{tab:seeds}. With real images, 58--67\% of discovery groups and 88--89\% of probe groups contain both outcomes at step 1 (means over steps 1--5: 0.63--0.66 and 0.60--0.62); with gray canvases, 2--5\% of discovery groups and no probe group do, and 95--100\% of groups are all wrong. Nonzero advantages nevertheless arise in 61--64\% of the gray-canvas discovery groups and 45--48\% of its probe groups at step 1: format varies within a group even when correctness does not. That source is short-lived. The nonzero-advantage fraction of the \cond{Gray}-trained runs falls to 0.26--0.32 over steps 1--5, their format reward reaches 0.999--1.000 by step 30 (Table~\ref{tab:seedrewards}), and their mean accuracy reward over steps 19--30 is 0.024--0.041, so by then an update occurs only in the rare group that contains a correct response.

\begin{table}[h]
\caption{Reward variation within response groups at the start of training, for the runs of Table~\ref{tab:seeds}, by training-row type (discovery, the identification probe, or find-and-bind for the named-target corpus). Columns: mean accuracy and format components at step 1; the fraction of groups containing both a correct and an incorrect response (``mixed'', the event of Eq.~(\ref{eq:pmix})); the fraction of groups with any within-group reward variation (``nonzero advantage''); the same two fractions averaged over steps 1--5; and the step-1 fractions of all-correct and all-wrong groups.}
\label{tab:learnsignal}
\centering
\scriptsize
\setlength{\tabcolsep}{3pt}
\begin{tabular}{l c l >{\raggedleft\arraybackslash}p{26pt} >{\raggedleft\arraybackslash}p{24pt} >{\raggedleft\arraybackslash}p{24pt} >{\raggedleft\arraybackslash}p{31pt} >{\raggedleft\arraybackslash}p{24pt} >{\raggedleft\arraybackslash}p{31pt} >{\raggedleft\arraybackslash}p{24pt} >{\raggedleft\arraybackslash}p{24pt}}
\toprule
condition & seed & rows & step-1 accuracy & step-1 format & mixed, step 1 & nonzero advantage, step 1 & mixed, steps 1--5 & nonzero advantage, steps 1--5 & all correct, step 1 & all wrong, step 1 \\
\midrule
\cond{Real} & 1 & discovery & 0.223 & 0.857 & 0.633 & 0.808 & 0.645 & 0.698 & 0.008 & 0.358 \\
\cond{Real} & 1 & probe & 0.435 & 0.875 & 0.892 & 0.933 & 0.612 & 0.622 & 0.025 & 0.083 \\
\cond{Real} & 2 & discovery & 0.178 & 0.883 & 0.575 & 0.800 & 0.632 & 0.690 & 0.008 & 0.417 \\
\cond{Real} & 2 & probe & 0.407 & 0.892 & 0.875 & 0.942 & 0.598 & 0.610 & 0.017 & 0.108 \\
\cond{Real} & 3 & discovery & 0.245 & 0.875 & 0.667 & 0.842 & 0.655 & 0.718 & 0.008 & 0.325 \\
\cond{Real} & 3 & probe & 0.428 & 0.890 & 0.892 & 0.933 & 0.622 & 0.637 & 0.025 & 0.083 \\
\cond{Gray}-trained & 1 & discovery & 0.003 & 0.862 & 0.017 & 0.608 & 0.025 & 0.265 & 0.000 & 0.983 \\
\cond{Gray}-trained & 1 & probe & 0.000 & 0.875 & 0.000 & 0.458 & 0.003 & 0.300 & 0.000 & 1.000 \\
\cond{Gray}-trained & 2 & discovery & 0.005 & 0.875 & 0.025 & 0.617 & 0.034 & 0.283 & 0.000 & 0.975 \\
\cond{Gray}-trained & 2 & probe & 0.000 & 0.885 & 0.000 & 0.475 & 0.002 & 0.298 & 0.000 & 1.000 \\
\cond{Gray}-trained & 3 & discovery & 0.010 & 0.867 & 0.050 & 0.642 & 0.021 & 0.255 & 0.000 & 0.950 \\
\cond{Gray}-trained & 3 & probe & 0.000 & 0.893 & 0.000 & 0.450 & 0.007 & 0.317 & 0.000 & 1.000 \\
named-target & 1 & find-and-bind & 0.155 & 0.867 & 0.517 & 0.875 & 0.547 & 0.635 & 0.000 & 0.483 \\
named-target & 1 & probe & 0.435 & 0.875 & 0.892 & 0.933 & 0.607 & 0.620 & 0.025 & 0.083 \\
named-target & 2 & find-and-bind & 0.137 & 0.873 & 0.475 & 0.842 & 0.512 & 0.628 & 0.000 & 0.525 \\
named-target & 2 & probe & 0.422 & 0.892 & 0.883 & 0.942 & 0.635 & 0.647 & 0.017 & 0.100 \\
named-target & 3 & find-and-bind & 0.183 & 0.883 & 0.608 & 0.900 & 0.585 & 0.675 & 0.000 & 0.392 \\
named-target & 3 & probe & 0.445 & 0.888 & 0.900 & 0.942 & 0.595 & 0.608 & 0.025 & 0.075 \\
\bottomrule
\end{tabular}
\par\vspace{2pt}
{\scriptsize\raggedright Within a seed, the \cond{Real} and named-target runs share the starting checkpoint, the probe rows and the sampling seed, so their step-1 probe rows coincide and diverge only afterwards.\par}
\end{table}

\subsection{Comparing audit measurements under a common success profile}\label{app:standardization}

Take the audit-joined rows of each condition's realized training stream, keeping their presentation multiplicities, and assign each row the stratum
\begin{equation}
s=\bigl(\min\{9,\lfloor 10\,b_i\rfloor\},\;\mathbf 1\{0<b_i<1\}\bigr).
\label{eq:strata}
\end{equation}
Let $\mathcal S$ be the strata represented in all four conditions, $n_{f,s}$ the row counts, and $\mu_{f,s}$ the mean $\Delta q_i$ within condition $f$ and stratum $s$. The common weights and the standardized mass are
\begin{equation}
w_s=\frac{\sum_f n_{f,s}}{\sum_{u\in\mathcal S}\sum_f n_{f,u}},
\qquad
\widetilde m(f)=\sum_{s\in\mathcal S}w_s\,\mu_{f,s}.
\label{eq:stdmass}
\end{equation}
Because $\Delta q_i=b_i-q_{\mathrm{blind},i}$, the same weights applied to the stratum means $\bar b_{f,s}$ and $\bar q_{f,s}$ of those rows give
\begin{equation}
\widetilde m(f)
=\underbrace{\sum_{s\in\mathcal S} w_s\,\bar b_{f,s}}_{\widetilde B(f)}
-\underbrace{\sum_{s\in\mathcal S} w_s\,\bar q_{f,s}}_{\widetilde Q(f)},
\label{eq:stddecomp}
\end{equation}
which separates the with-image and blind components of the standardized audit; within the coarse bins, $\widetilde B(f)$ need not be identical across conditions.

Common stratum weights place the audit measurements on a shared initial-success profile. Each outcome remains the measured held-out accuracy of its original training run; the standardization changes the audit summary, not the training distribution. Table~\ref{tab:slopes} reports the resulting associations and the separate equal-exposure comparisons. The slopes are ordinary least-squares slopes with an intercept over the four conditions, of step-30 held-out accuracy on $m(f)$ or on $\widetilde m(f)$; Appendix~\ref{app:stats} describes their bootstrap intervals, which resample held-out scene programs paired across conditions with the audit measurements held fixed.

\clearpage
\section{The Experimental Program at a Glance}\label{app:program}
\begin{figure}[!h]
\centering
\includegraphics[width=\linewidth]{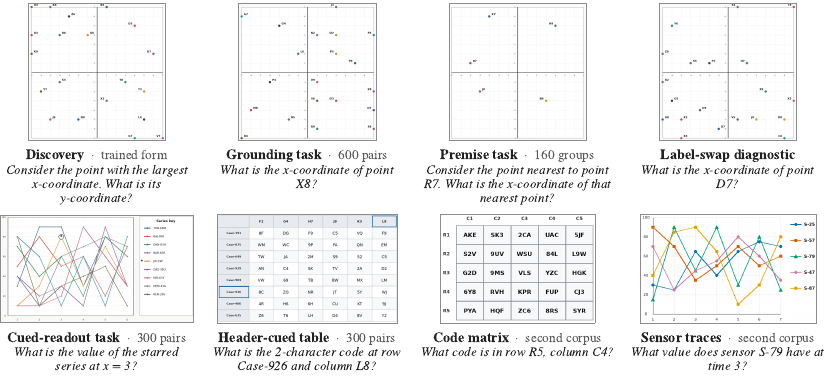}
\caption{Eight of the nine task families: coordinate scenes (top) and other formats (bottom).}
\label{fig:gallery}
\end{figure}

\setlength{\intextsep}{8pt plus 2pt minus 2pt}
The coordinate scenes of \S\ref{sec:make} are this work's controlled training testbed; the wider program spans nine counterfactual task families, two public training corpora, seven audited benchmarks, four models from three families, and over thirty GRPO runs of up to 400 steps (Figure~\ref{fig:gallery}, Table~\ref{tab:program}).
\begin{table}[!h]
\caption{The experimental program in numbers.}
\label{tab:program}
\centering
\scriptsize
\setlength{\tabcolsep}{3pt}
\begin{tabular}{>{\raggedright\arraybackslash}p{2.8cm} >{\raggedright\arraybackslash}p{10.7cm}}
\toprule
component & scale \\
\midrule
constructed evaluations & a 1,200-pair suite and its regenerated 1,200-pair twin; discovery scenes at 8, 12 and 20 points, with 450 confirmatory scene programs; an untouched replication set of 333 scene programs per pair role (Table~\ref{tab:seeds}); 160 premise groups (an untrained nearest-point relation task); 100 label-swap diagnostic pairs \\
constructed training data & 2,880 discovery and probe prompts; a second corpus of 3,000 pairs and 300 answer-preserving twins \\
public image corpora & Geometry3K (1,288 training, 601 test items); ViRL39K (23,542 training, 4,239 held-out items) \\
public benchmarks & seven, more than 10,000 items, audited with and without the image at 3B and 7B (3B: Table~\ref{tab:audit}) \\
models & Qwen2.5-VL-3B and 7B trained; Gemma-3 and InternVL3-9B audited (Table~\ref{tab:crossfamily}) \\
training runs & over thirty GRPO runs of 30 to 400 steps, one to five runs per condition (main-text runs: Table~\ref{tab:recipes}) \\
base-model audit & 6,976 items of the constructed corpus and ViRL39K, each with 16 responses with and 16 without the image \\
\bottomrule
\end{tabular}
\end{table}

\textbf{More than one kind of question.} The grounding task and its twin predate the training corpus; at step 30, $+0.160$ (95\% CI [0.130, 0.190]) of its gain is attributable to training images (\cond{Gray}-trained control: $+0.002$). The premise task asks for a nearest-point relation that was never trained: at step 30, final-answer pair accuracy rises from 19 to 33 of 80 pairs at 5 points and from 22 to 28 at 20 points, and is zero under a gray canvas (no \cond{Gray}-trained control). On the second corpus (3B, 120 steps, three templates), four configurations improved cued readout ($+0.15$ to $+0.23$) but not the grounding task.

\textbf{Public images, in training and at test.} We trained on both public corpora under four access conditions (ViRL39K, 4,239 held-out items, each condition evaluated as trained: \cond{Real} $+0.204$, \cond{Gray} $+0.149$). Constructed data was also mixed into ViRL39K training: at $f=\tfrac13$, 160 of every 240 prompts are ViRL39K; never-trained discovery gains $+0.275$ against $-0.010$ for the all-ViRL39K run, and every mixture's held-out ViRL39K accuracy is at least 0.20 above its image-removed accuracy (\S\ref{sec:dose}).

\textbf{Families, scales and horizons.} Discovery performance improves at both trained scales (3B: $+0.300$ to $+0.535$ in 30 steps; Table~\ref{tab:capacity}). Two families we did not train, Gemma-3 and InternVL3-9B, score 0.333 and 0.681 on the grounding task, match it on the twin within 0.005, and fall to exactly 0.000 without images. Ordinary training ran 400 steps in two runs (\S\ref{sec:decay}); constructed-corpus training ran 100 segmented steps in three runs (two of seed 0, one of seed 1), long after reward saturates at step 13: the grounding task reads 0.948, 0.958, 0.948 at steps 50, 75, 100 (second run 0.960, 0.968, 0.962; seed 1 at step 100, 0.942).\label{end:appK}
\setlength{\intextsep}{10pt plus 2pt minus 2pt}

\clearpage
\AppendixBarrier
\section{Extended Related Work}\label{app:related}
\hyphenation{ViCrit}
\textbf{RLVR and what it changes.} Tulu 3 named reinforcement learning with verifiable rewards (RLVR), which replaces a learned reward model with a verification function~\citep{lambert2025tulu3}, and DeepSeek-R1-Zero optimizes rule-based accuracy and format rewards with GRPO~\citep{guo2025deepseekr1}. Vision-R1, MM-Eureka, and VL-Rethinker extend RLVR to visual reasoning~\citep{huang2026visionr1,meng2025mmeureka,wang2025vlrethinker}. Whether RLVR develops new skills or elicits existing ones is open: pass@$k$ analyses find that base models can match or exceed RL-trained ones as sampling budgets grow~\citep{yue2025does}, and some Qwen models improve at mathematics under random or incorrect rewards~\citep{shao2026spurious}. Process supervision instead scores each reasoning step: a reward model trained on step-level human feedback reranks sampled solutions to mathematics problems, without RL~\citep{lightman2024verify}. Our visual-access comparisons hold the verifier and training setup fixed within each corpus: blind training recovers much of the real-image gain on Geometry3K (\S\ref{sec:blind}) but, at matched step 30, essentially none of the discovery gain from training on visually resolvable tasks (\S\ref{sec:drives}).

\textbf{Visual dependence in evaluation.} Controlled contrasts expose shortcuts that sustain accuracy without the intended computation~\citep{geirhos2020shortcut}. VQA v2 counters language priors with complementary images~\citep{goyal2017making}; VQA-CP changes the answer distribution of each question type between training and test while keeping the images' distribution, and existing models degrade sharply~\citep{agrawal2018vqacp}; MMVP probes visual differences between images with similar CLIP representations~\citep{tong2024eyes}; and BLINK targets perception-intensive tasks~\citep{fu2024blink}. MMStar filters out text-answerable questions and evaluates the same vision-language model with and without images, while its leakage metric measures image-free improvement over the language-model backbone~\citep{chen2024mmstar}; MMMU-Pro also filters out text-answerable questions~\citep{yue2025mmmupro}. Blind gains probe the training stage instead: visual information is withheld during RL and restored at evaluation, so an improvement can depend on the image at test without having been learned from images. Our benchmark audit measures dependence by chance-corrected blind retention (\S\ref{sec:necessity}), and programmatic pairs with controlled target cues identify which operation training improves (\S\ref{sec:ladder}).

\textbf{Paired scoring.} MMVP scores a CLIP-blind pair only when both of its questions are answered correctly~\citep{tong2024eyes}; NaturalBench pairs each question with two images that yield different answers, and its question accuracy awards a point only when the question is answered correctly for both images~\citep{li2024naturalbench}. The pairs of our counterfactual suite are generated programmatically: the question stays fixed and one visual fact changes so that the two correct answers differ. Eq.~(\ref{eq:pairacc}) applies the same rule to these pairs, and because their accepted answers are disjoint, an image-independent greedy predictor scores exactly zero (Proposition~\ref{prop:pairzero}). The guarantee concerns pair accuracy; for discovery and its identification probe, scored per question, accuracy without the image is measured directly.

\textbf{Visual skill under reward optimization.} Multimodal reasoning models hallucinate more than their non-reasoning counterparts on perception tasks, and longer reasoning chains reduce attention to visual inputs~\citep{xu2025morethinking}. Reasoning-only RLVR on Qwen2.5-VL-7B lowers an out-of-domain segmentation score below the base model's while mathematics scores rise~\citep{phan2026beyond}; VOLD, which transfers reasoning to a vision-language model with text-only training, reports a small perception decrease that an image-trained reasoning model shares and attributes it to reasoning-focused training rather than to text-only data~\citep{bousselham2026vold}; and after RLVR on Geometry3K with only accuracy and format rewards, perception on 50 MathVista problems does not differ significantly from the base model's~\citep{xiao2026perceptionr1}. Grounding corrosion (\S\ref{sec:corrosion}) is a training-time deterioration measured on counterfactual pairs: under prolonged real-image training with a trainable vision encoder, grounding pair accuracy falls below the base model's while benchmark gains persist. In a separate analysis, independent blind-trained runs lose many of the same previously solved pairs, usually with identical wrong answers that real-image-trained runs often repeat. Aggregate scores show the decline; recurring errors show which examples independent runs lose and which wrong answers they share.

\textbf{Perception-aware reinforcement learning.} PAPO adds to the objective an implicit perception loss that maximizes the KL divergence between response distributions under original and masked images, leaving the reward unchanged~\citep{wang2026papo}; Visionary-R1 imposes a caption--reason--answer structure and rewards captions from which the question can be answered~\citep{xia2026visionaryr1}; DeepEyes rewards image-cropping tool use when the answer is correct and, for its fine-grained perception data, keeps examples that crops make solvable~\citep{zheng2026deepeyes}. Perception-R1 adds a visual perception reward, judged by a language model, to the accuracy and format rewards~\citep{xiao2026perceptionr1}, and Visual-RFT trains on perception tasks with task-specific verifiable rewards, such as IoU for object detection~\citep{liu2025visualrft}. A second method also named Perception-R1 trains perception policies with GRPO and rule-based rewards (IoU for grounding, Euclidean distance for counting, edit distance for OCR) and observes that perceptual perplexity is a major factor in RL's effectiveness~\citep{yu2025perceptionr1}. These methods act through the objective, the response structure, added evidence, or the reward, and the last two also through the task. We use standard GRPO with the correctness-and-format reward fixed within the constructed-corpus comparisons. Task design controls the visual fact relevant to the answer and the target information supplied (\S\ref{sec:shape}). A matched blind-trained control, with the same prompts, starting checkpoint, reward, and budget, then identifies how much of the gain comes from training-time visual information (\S\ref{sec:drives}).

\textbf{Learnability, visual necessity, and data selection.} DAPO's dynamic sampling keeps prompts with both successful and unsuccessful responses~\citep{yu2025dapo}; ThinkLite-VL selects data by the tree-search iterations a problem needs~\citep{wang2025thinklite}. Both measure how hard a problem is for the model. Our learnability band, $0<b_i<1$, shares this concern; it is measured once on the base model and describes the corpus rather than filtering it (\S\ref{sec:measure}). VisNec uses the name ``visual necessity'' for the loss reduction the image provides under teacher forcing and selects instruction-tuning data with it, without RL~\citep{dong2026visnec}; our $\Delta q_i$ is a difference of sampled correct-answer rates with and without the image. This measure adds what DAPO and ThinkLite-VL lack: how much the image raises the base model's probability of a correct answer, which characterizes the training mixtures whose composition shapes acquisition (\S\ref{sec:dose}). Difficulty-based selection asks how hard a task is; resolvability also asks what evidence and operation successful answering require.\label{end:appL}

Two verifiable vision-centric training tasks are the closest precedents for task construction. ViCrit~\citep{wang2025vicrit} injects a small error into a human-written caption and rewards locating the corrupted span given the image, on the grounds that vision-centric tasks that are both challenging and verifiable are scarce; Visual Jigsaw~\citep{wu2026visualjigsaw} shuffles image patches, video clips or 3D views and rewards recovering the order, a self-supervised task that needs the image by construction. Both build tasks whose reward requires perception and report benchmark gains. Neither measures whether the gains are learned from the images (the matched blind-training control of \S\ref{sec:drives}), which visual operation improved (the cue hierarchy of \S\ref{sec:ladder}), or what the same reward teaches when the task admits a shortcut (the shortcut-accessible twin of \S\ref{sec:drives}); those three measurements are what this paper adds. VisNec uses ``visual necessity'' for the reduction in teacher-forced loss supplied by the image and selects instruction-tuning data with it~\citep{dong2026visnec}; our $\Delta q_i$ measures the change in sampled correctness under the RL verifier.

\end{document}